\documentclass[10pt,twocolumn,letterpaper]{article}

\usepackage[pagenumbers]{cvpr} % To force page numbers, e.g. for an arXiv version

\usepackage{graphicx}
\usepackage{amsmath}
\usepackage{amssymb}
\usepackage{booktabs}

\usepackage{enumitem}
\usepackage[dvipsnames]{xcolor}
\usepackage{mathtools}
\usepackage[linesnumbered,ruled,vlined]{algorithm2e}
\usepackage{floatrow}
\usepackage{multirow}
\usepackage{makecell}
\usepackage{etoolbox}
\usepackage{stfloats}

\graphicspath{{resources/}}

\SetKwInOut{KwIn}{Input}
\SetKwInOut{KwOut}{Output}
\SetKw{KwAnd}{and}

\SetCommentSty{mycommfont}
\SetAlgoNlRelativeSize{0}

\let\originalleft\left
\let\originalright\right
\renewcommand{\left}{\mathopen{}\mathclose\bgroup\originalleft}
\renewcommand{\right}{\aftergroup\egroup\originalright}
\DeclareMathOperator*{\argmax}{arg\,max}
\DeclareMathOperator*{\argmin}{arg\,min}
\newcommand{\diff}{\mathop{}\!\mathrm{d}}
\let\originalpartial\partial
\renewcommand{\partial}{\mathop{}\!\mathrm{\originalpartial}}
\newcommand{\expect}{\mathop{\mathbb{E}}}
\newcommand{\diag}{\mathop{\mathrm{diag}}}

\newfloatcommand{capbtabbox}{table}[][\FBwidth]

\makeatletter
\renewcommand{\paragraph}{
  \@startsection{paragraph}{4}
  {\z@}{1.0ex \@plus 0ex \@minus .3ex}{-1em}
  {\normalfont\normalsize\bfseries}
}
\makeatother

\makeatletter
\patchcmd{\@algocf@start}% <cmd>
  {-1.5em}% <search>
  {-0.5em}% <replace>
  {}{}% <success><failure>
\makeatother

\usepackage[pagebackref,breaklinks,colorlinks]{hyperref}

\usepackage[capitalize]{cleveref}
\crefname{section}{Sec.}{Secs.}
\Crefname{section}{Section}{Sections}
\Crefname{table}{Table}{Tables}
\crefname{table}{Tab.}{Tabs.}

\def\cvprPaperID{1659} % *** Enter the CVPR Paper ID here
\def\confName{CVPR}
\def\confYear{2022}

\begin{document}


\title{EPro-PnP: Generalized End-to-End Probabilistic Perspective-n-Points \\
for Monocular Object Pose Estimation}

\author{
Hansheng Chen,\negmedspace\textsuperscript{1,2,\textasteriskcentered}
Pichao Wang,\negmedspace\textsuperscript{2,\textdagger}
Fan Wang,\negmedspace\textsuperscript{2}
Wei Tian,\negmedspace\textsuperscript{1,\textdagger}
Lu Xiong,\negmedspace\textsuperscript{1}
Hao Li\textsuperscript{2} \\
\textsuperscript{1}School of Automotive Studies, Tongji University \qquad \textsuperscript{2}Alibaba Group \\
{\tt\small
hanshengchen97@gmail.com \{tian\_wei, xiong\_lu\}@tongji.edu.cn
} \\
{\tt\small
\{pichao.wang, fan.w, lihao.lh\}@alibaba-inc.com
}
}

\maketitle

\begingroup
\renewcommand{\thefootnote}{\fnsymbol{footnote}}
\footnotetext[1]{Part of work done during an internship at Alibaba Group.}
\footnotetext[2]{Corresponding authors: Pichao Wang, Wei Tian.}
\endgroup

%%%%%%%%% ABSTRACT
\begin{abstract}

\setcounter{footnote}{2}

Locating 3D objects from a single RGB image via Perspective-n-Points (PnP) is a long-standing problem in computer vision. Driven by end-to-end deep learning, recent studies suggest interpreting PnP as a differentiable layer, so that 2D-3D point correspondences can be partly learned by backpropagating the gradient \wrt object pose. Yet, learning the entire set of unrestricted 2D-3D points from scratch fails to converge with existing approaches, since the deterministic pose is inherently non-differentiable.
In this paper, we propose the EPro-PnP, a probabilistic PnP layer for general end-to-end pose estimation, which outputs a distribution of pose on the SE(3) manifold, essentially bringing categorical Softmax to the continuous domain.
The 2D-3D coordinates and corresponding weights are treated as intermediate variables learned by minimizing the KL divergence between the predicted and target pose distribution.
The underlying principle unifies the existing approaches and resembles the attention mechanism.
% Without bells and whistles, 
EPro-PnP significantly outperforms competitive baselines, closing the gap between PnP-based method and the task-specific leaders on the LineMOD 6DoF pose estimation and nuScenes 3D object detection benchmarks.\footnote{Code: \url{https://github.com/tjiiv-cprg/EPro-PnP}}
   
\end{abstract}

%%%%%%%%% BODY TEXT
\section{Introduction}

\begin{figure}[t]
\begin{center}
  \includegraphics[width=0.9\linewidth]{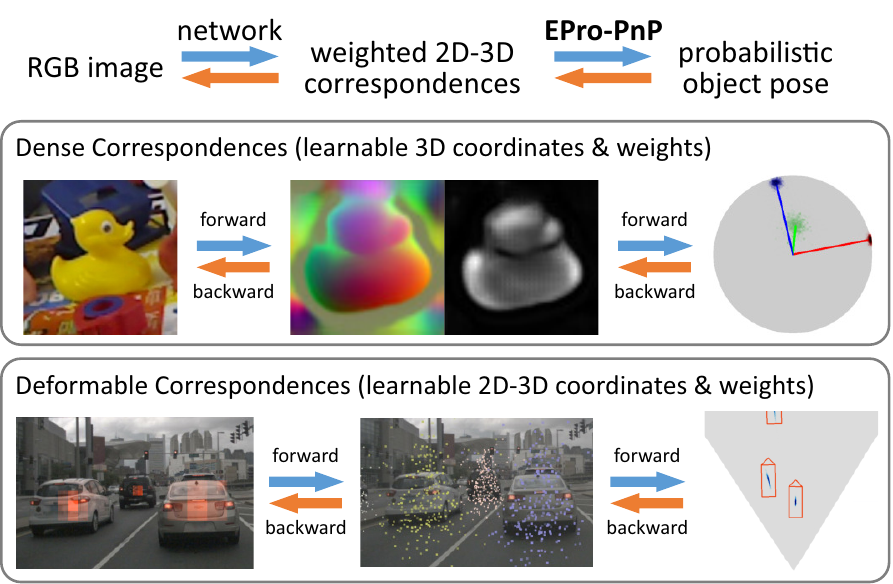}
\end{center}
\vspace{-0.5ex}
\caption{EPro-PnP is a general solution to end-to-end 2D-3D correspondence learning. In this paper, we present two distinct networks trained with EPro-PnP: (a) an off-the-shelf dense correspondence network whose potential is unleashed by end-to-end training, (b) a novel deformable correspondence network that explores new possibilities of fully learnable 2D-3D points.}
\label{fig:header}
\end{figure}

Estimating the pose (\ie, position and orientation) of 3D objects from a single RGB image is an important task in computer vision.
This field is often subdivided into specific tasks, \eg, 6DoF pose estimation for robot manipulation and 3D object detection for autonomous driving. Although they share the same fundamentals of pose estimation, the different nature of the data leads to biased choice of methods. Top performers~\cite{pgd, dd3d, detr3d} on the 3D object detection benchmarks~\cite{nuscenes,kitti} fall into the category of direct 4DoF pose prediction, leveraging the advances in end-to-end deep learning. On the other hand, the 6DoF pose estimation benchmark~\cite{linemod} is largely dominated by geometry-based methods~\cite{repose, DPOD}, which exploit the provided 3D object models and achieve a stable generalization performance. However, it is quite challenging to bring together the best of both worlds, \ie, training a geometric model to learn the object pose in an end-to-end manner.

There has been recent proposals for an end-to-end framework based on the Perspective-n-Points (PnP) approach~\cite{dsac, dsac++, blindpnp, BPnP}. The PnP algorithm itself solves the pose from a set of 3D points in object space and their corresponding 2D projections in image space, leaving the problem of constructing these correspondences.
Vanilla correspondence learning~\cite{pvnet, hybridpose, bb8, NOCS, pix2pose, DPOD, CDPN, RTM3D, deep-manta, pix2pose} leverages the geometric prior to build surrogate loss functions, forcing the network to learn a set of pre-defined correspondences. End-to-end correspondence learning~\cite{dsac, dsac++, blindpnp, BPnP} interprets the PnP as a differentiable layer and employs pose-driven loss function, so that gradient of the pose error can be backpropagated to the 2D-3D correspondences.

However, existing work on differentiable PnP learns only a portion of the correspondences (either 2D coordinates~\cite{BPnP}, 3D coordinates~\cite{dsac, dsac++} or corresponding weights~\cite{blindpnp}), assuming other components are given \emph{a priori}. This raises an important question: 
why not learn the entire set of points and weights altogether in an end-to-end manner?
% by backpropagating through the differentiable PnP
The simple answer is: the solution of the PnP problem is inherently non-differentiable at some points, causing training difficulties and convergence issues. For example, a PnP problem can have ambiguous solutions~\cite{manhardt2019, Schweighofer2006}.
% This is also the reason why current end-to-end approaches~\cite{dsac, dsac++,BPnP,blindpnp} are still coupled with surrogate loss (\eg, direct regression or reprojection loss) that helps convergence.

To overcome the above limitations, we propose a generalized \textbf{e}nd-to-end \textbf{pro}babilistic \textbf{PnP} (EPro-PnP) approach that enables learning the weighted 2D-3D point correspondences entirely from scratch (Figure~\ref{fig:header}). The main idea is straightforward: deterministic pose is non-differentiable, but the probability density of pose is apparently differentiable, just like categorical classification scores. Therefore, we interpret the output of PnP as a probabilistic distribution parameterized by the learnable 2D-3D correspondences. During training, the Kullback-Leibler (KL) divergence between the predicted and target pose distributions is minimized as the loss function, which can be efficiently implemented by the Adaptive Multiple Importance Sampling~\cite{amis} algorithm.

As a general approach, EPro-PnP inherently unifies existing correspondence learning techniques (Section~\ref{overview}). Moreover, just like the attention mechanism~\cite{vaswani2017attention}, the corresponding weights can be trained to automatically focus on important point pairs, allowing the networks to be designed with inspiration from attention-related work~\cite{detr,nonlocal,deformabledetr}.
% We therefore design the networks with inspiration from dense~\cite{detr, nonlocal} and deformable~\cite{deformabledetr} attention in object detection.
% , which fits dense correspondence~\cite{pix2pose, CDPN, DPOD, monorun} and a novel form of deformable correspondence, respectively.

To summarize, our main contributions are as follows: 
\begin{itemize}[noitemsep,topsep=0.7ex,partopsep=0.7ex]
    \item We propose the EPro-PnP, a probabilistic PnP layer for general end-to-end pose estimation via learnable 2D-3D correspondences. 
    \item We demonstrate that EPro-PnP can easily reach top-tier performance for 6DoF pose estimation by simply inserting it into the CDPN~\cite{CDPN} framework.
    \item We demonstrate the flexibility of EPro-PnP by proposing \emph{deformable correspondence learning} for accurate 3D object detection, where the entire 2D-3D correspondences are learned from scratch.
\end{itemize}

% summary & list contribution

\section{Related Work}

\paragraph{Geometry-Based Object Pose Estimation}
In general, geometry-based methods exploit the points, edges or other types of representation that are subject to the projection constraints under the perspective camera. Then, the pose can be solved by optimization. A large body of work utilizes point representation, which can be categorized into sparse keypoints and dense correspondences. BB8~\cite{bb8} and RTM3D~\cite{RTM3D} locate the corners of the 3D bounding box as keypoints, while PVNet~\cite{pvnet} defines the keypoints by farthest point sampling and Deep MANTA~\cite{deep-manta} by handcrafted templates. On the other hand, dense correspondence methods \cite{pix2pose, CDPN, DPOD, monorun, NOCS} predict pixel-wise 3D coordinates within a cropped 2D region.
% Moreover, HybridPose~\cite{hybridpose} adopts a hybrid representation consisting of keypoints, edge vectors, and symmetry correspondences.
Most existing geometry-based methods follow a two-stage strategy, where the intermediate representations (\ie, 2D-3D correspondences) are learned with a surrogate loss function, which is sub-optimal compared to end-to-end learning.

\paragraph{End-to-End Correspondence Learning}

To mitigate the limitation of surrogate correspondence learning, end-to-end approaches have been proposed to backpropagate the gradient from pose to intermediate representation. By differentiating the PnP operation, Brachmann and Rother~\cite{dsac++} propose a dense correspondence network where 3D points are learnable, BPnP~\cite{BPnP} predicts 2D keypoint locations, and BlindPnP~\cite{blindpnp} learns the corresponding weight matrix given a set of unordered 2D/3D points. Beyond point correspondence, RePOSE~\cite{repose} proposes a feature-metric correspondence network trained in a similar end-to-end fashion. The above methods are all coupled with surrogate regularization loss, otherwise convergence is not guaranteed due to the non-differentiable nature of deterministic pose. Under the probabilistic framework, these methods can be regarded as a Laplace approximation approach (Section~\ref{overview}) or a local regularization technique (Section~\ref{localreg}).

\paragraph{Probabilistic Deep Learning}
Probabilistic methods account for uncertainty in the model and the data, known respectively as epistemic and aleatoric uncertainty~\cite{kendall2017uncertainties}. The latter involves interpreting the prediction as learnable probabilistic distributions. Discrete categorical distribution via Softmax has been widely adopted as a smooth approximation of one-hot $\argmax$ for end-to-end classification. This inspired works such as DSAC~\cite{dsac}, a smooth RANSAC with a finite hypothesis pool. Meanwhile, tractable parametric distributions (\eg, normal distribution) are often used in predicting continuous variables~\cite{klloss, wu2020unsupervised, kendall2017uncertainties, VAE, Gilitschenski2020, monorun}, and mixture distributions can be employed to further capture ambiguity~\cite{makansi2019, Bishop94mixturedensity, Brachmann_2016_CVPR}, \eg, ambiguous 6DoF pose~\cite{bui20206d}. In this paper, we propose yet a unique contribution: backpropagating a complicated continuous distribution derived from a nested optimization layer (the PnP layer), essentially making it a continuous counterpart of Softmax.
% In this paper, the aleatoric uncertainty of object pose is characterized by the PnP model. Similar to DSAC~\cite{dsac}, PnP is made truly differentiable by converting the deterministic $\argmin$ optimum into a continuous probabilistic representation.

\section{Generalized End-to-End Probabilistic PnP}

\subsection{Overview} \label{overview}

Given an object proposal,
% from a base detector, be it a 2D bounding box or a feature vector, 
our goal is to predict a set $X = \left\{x^\text{3D}_i,x^\text{2D}_i,w^\text{2D}_i\,\middle|\,i=1\cdots N\right\}$ of $N$ corresponding points, with 3D object coordinates $x^\text{3D}_i \in \mathbb{R}^3$, 2D image coordinates $x^\text{2D}_i \in \mathbb{R}^2$, and 2D weights $w^\text{2D}_i \in \mathbb{R}^2_+ $, from which a weighted PnP problem can be formulated to estimate the object pose relative to the camera.

The essence of a PnP layer is searching for an optimal pose $y$ (expanded as rotation matrix $R$ and translation vector $t$) that minimizes the cumulative squared weighted reprojection error:
\vspace{-1ex}
\begin{equation}
\smash[b]{\argmin_{y} \frac{1}{2} \sum_{i=1}^N \left\| \smash[b]{\underbrace{w_i^\text{2D} \circ \left( \pi(Rx_i^\text{3D} + t) - x_i^\text{2D} \right)}_{f_i(y) \in \mathbb{R}^2}} \right\|^2,}
\vphantom{\underbrace{\left((O_o^O)\right)}_{o}}
\label{eqn:basicpnp}
\end{equation}
where $\pi(\cdot)$ is the projection function with camera intrinsics involved, $\circ$ stands for element-wise product, and $f_i(y)$ compactly denotes the weighted reprojection error.

Eq.~(\ref{eqn:basicpnp}) formulates a non-linear least squares problem that may have non-unique solutions, \ie, pose ambiguity~\cite{manhardt2019, Schweighofer2006}. Previous work~\cite{dsac++, BPnP, blindpnp} only backpropagates through a local solution $y^\ast$, which is inherently unstable and non-differentiable.
% where the $\argmin$ function is not continuous, let alone differentiable. The continuity breaks typically when the global optimum switches from one mode to another, due to pose ambiguity~\cite{manhardt2019, Schweighofer2006}. 
% Previous efforts~\cite{dsac++, BPnP, blindpnp} on backwarding through the PnP operation only derives a single local optimum, which does not guarantee global convergence.
To construct a differentiable alternative for end-to-end learning, we model the PnP output as a distribution of pose, which guarantees differentiable probability density. Consider the cumulative error to be the negative logarithm of the likelihood function $p(X|y)$ defined as:
\begin{equation}
p \left(X \middle| y \right) = \exp -\frac{1}{2} \sum_{i=1}^N \left\| f_i(y) \right\|^2 .
\label{nll}
\end{equation}
With an additional prior pose distribution $p(y)$, we can derive the posterior pose $p(y|X)$ via the Bayes theorem. Using an \emph{uninformative prior}, the posterior density is simplified to the normalized likelihood:
\begin{equation}
p(y|X)
% = \frac{p(X|y)}{ \int p(X|y) \diff{y}}
= \frac{\exp -\frac{1}{2} \sum_{i=1}^N \left\| f_i(y) \right\|^2}{ \int \exp -\frac{1}{2} \sum_{i=1}^N \left\| f_i(y) \right\|^2 \diff{y}}.
\label{posterior}
\end{equation}
Eq.~(\ref{posterior}) can be interpreted as a continuous counterpart of categorical Softmax.

\paragraph{KL Loss Function}
During training, given a target pose distribution with probability density $t(y)$, the KL divergence $D_\text{KL}\left(t(y) \| p(y|X) \right)$ is minimized as training loss. Intuitively, pose ambiguity can be captured by the multiple modes of $p(y|X)$, and convergence is ensured such that wrong modes are suppressed by the loss function. Dropping the constant, the KL divergence loss can be written as:
\begin{equation}
L_\text{KL} = -\int t(y) \log p(X|y) \diff{y} + \log \int p(X|y) \diff{y}.
\end{equation}
We empirically found it effective to set a narrow (Dirac-like) target distribution centered at the ground truth $y_\text{gt}$, yielding the simplified loss (after substituting Eq.~(\ref{nll})):
% \vspace{-1ex}
\begin{equation} 
L_\text{KL} = \smash[b]{\underbrace{\frac{1}{2} \sum_{i=1}^N \left\| f_i(y_\text{gt}) \right\|^2}_{\mathclap{L_\text{tgt} \text{ (reproj. at target pose)}}} + \underbrace{\log \int \exp - \frac{1}{2} \sum_{i=1}^N \left\| f_i(y) \right\|^2 \diff{y}}_{L_\text{pred}\text{ (reproj. at predicted pose)}}}. \vphantom{\underbrace{\sum}_{O_o}}
\label{symbloss}
\end{equation}
The only remaining problem is the integration in the second term, which is elaborated in Section~\ref{mcloss}.

\begin{figure}[t]
\begin{center}
    \includegraphics[width=1.0\linewidth]{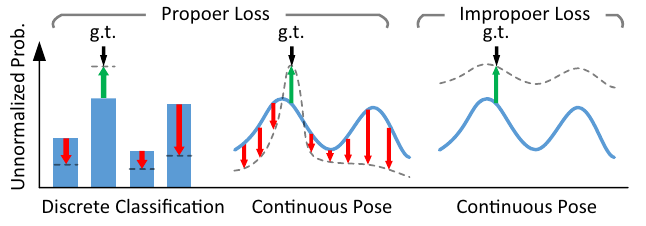}
\end{center}
\vspace{-1ex}
\caption{\textbf{Learning a discrete classifier \vs Learning the continuous pose distribution}. A discriminative loss function (left) shall encourage the unnormalized probability for the correct prediction as well as penalize for the incorrect. 
A one-sided loss (right) will degrade the distribution if the model is not well-regularized.}
\label{fig:distr}
\end{figure}

\paragraph{Comparison to Reprojection-Based Method}
The two terms in Eq.~(\ref{symbloss}) are concerned with the reprojection errors at target and predicted pose respectively. The former is often used as a surrogate loss in previous work~\cite{BPnP, dsac++, monorun}. However, the first term alone cannot handle learning all 2D-3D points without imposing strict regularization, as the minimization could simply drive all the points to a concentrated location without pose discrimination. The second term originates from the normalization factor in Eq.~(\ref{posterior}), and is crucial to a discriminative loss function, as shown in Figure~\ref{fig:distr}.

\paragraph{Comparison to Implicit Differentiation Method} 
Existing work on end-to-end PnP~\cite{BPnP, blindpnp}
% backpropagates the gradient \wrt $y^\ast$ by
derives a single solution of a particular solver $y^\ast = \mathit{PnP}(X)$ via implicit function theorem~\cite{declarative}. In the probabilistic framework, this is essentially the Laplace method that approximates the posterior by $\mathcal{N}(y^\ast, \Sigma_{y^\ast})$, where both $y^\ast$ and $\Sigma_{y^\ast}$ can be estimated by the PnP solver with analytical derivatives~\cite{monorun}. 
% Therefore, the KL divergence loss is analytically tractable.
A special case is that, with $\Sigma_{y^\ast}$ simplified to be isotropic, the approximated KL divergence can be simplified to the L2 loss $\|y^\ast - y_\text{gt}\|^2$ used in \cite{blindpnp}. However, the Laplace approximation is inaccurate for non-normal posteriors with ambiguity, therefore does not guarantee global convergence.

\subsection{Monte Carlo Pose Loss} \label{mcloss}

In this section, we introduce a GPU-friendly efficient Monte Carlo approach to the integration in the proposed loss function, based on the Adaptive Multiple Importance Sampling (AMIS) algorithm~\cite{amis}.
% , which can be easily implemented for efficient parallel computation on GPU.

Considering $q(y)$ to be the probability density function of a proposal distribution that approximates the shape of the integrand $\exp -\frac{1}{2} \sum_{i=1}^N \left\| f_i(y) \right\|^2$, and $y_j$ to be one of the $K$ samples drawn from $q(y)$, the estimation of the second term $L_\text{pred}$ in Eq.~(\ref{symbloss}) is thus:
\begin{equation}
 L_\text{pred} \approx \log \frac{1}{K} \sum_{j=1}^K \underbrace{\frac{\exp - \frac{1}{2} \sum_{i=1}^N \left\| f_i(y_j)\right\|^2}{q(y_j)}}_{v_j \text{ (importance weight)}},
\label{vanillais}
\end{equation}
where $v_j$ compactly denotes the importance weight at $y_j$. Eq.~(\ref{vanillais}) gives the vanilla importance sampling, where the choice of proposal $q(y)$ strongly affects the numerical stability. The AMIS algorithm is a better alternative as it iteratively adapts the proposal to the integrand. 

In brief, AMIS utilizes the sampled importance weights from past iterations to estimate the new proposal. Then, all previous samples are re-weighted as being homogeneously sampled from a mixture of the overall sum of proposals. Initial proposal can be determined by the mode and covariance of the predicted pose distribution (see supplementary for details). A pseudo-code is given in Algorithm~\ref{amisalg}.

\paragraph{Choice of Proposal Distribution} 
The proposal distributions for position and orientation have to be chosen separately in a decoupled manner, since the orientation space is non-Euclidean. For position, we adopt the 3DoF multivariate t-distribution.
% as in the original AMIS paper.
For 1D yaw-only orientation, we use a mixture of von Mises and uniform distribution. For 3D orientation represented by unit quaternion, the angular central Gaussian distribution~\cite{ACG} is adopted. 
% Details on the parameter estimation are given in the supplementary materials.

\subsection{Backpropagation} \label{backprop}

% To intuitively understand how a network actually learns $X$ via the proposed EPro-PnP, the derivatives of the loss function are derived in this section for further analysis.

In general, the partial derivatives of the loss function defined in Eq.~(\ref{symbloss}) is:
\vspace{-1ex}
\begin{equation}
\smash[b]{\frac{\partial{L_\text{KL}}}{\partial{(\cdot)}} = \frac{\originalpartial}{\partial{(\cdot)}} \frac{1}{2} \sum_{i=1}^N \left\| f_i(y_\text{gt}) \right\|^2 -
\hspace{-2ex} \expect_{y \sim p(y|X)}{\frac{\originalpartial}{\partial{(\cdot)}} \frac{1}{2} \sum_{i=1}^N \left\| f_i(y) \right\|^2}\mspace{-7mu},} \vphantom{\frac{}{O_o}}
\label{derivative}
\end{equation}
where the first term is the gradient of reprojection errors at target pose, and the second term is the expected gradient of reprojection errors over predicted pose distribution, which is approximated by backpropagating each weighted sample in the Monte Carlo pose loss.

\begin{figure}[t]
\vspace{-1.0ex}
\begin{center}
    \includegraphics[width=0.89\linewidth]{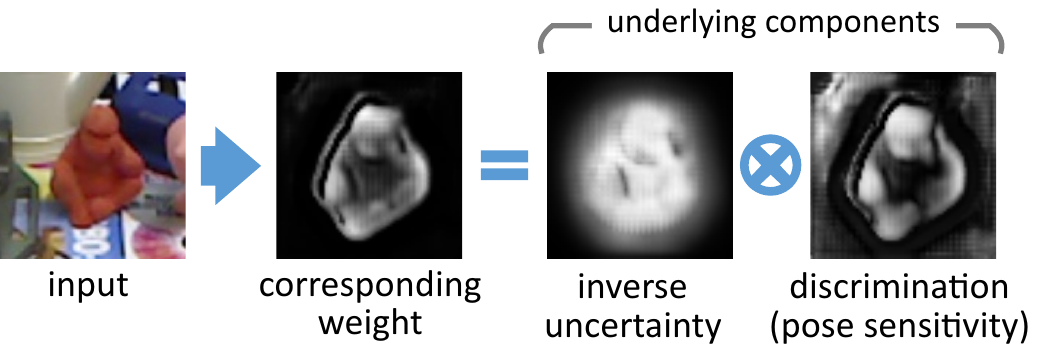}
\end{center}
\vspace{-1.2ex}
\caption{\textbf{The learned corresponding weight} can be factorized into inverse uncertainty and discrimination. 
Typically, inverse uncertainty roughly resembles the foreground mask, while discrimination emphasizes the 3D extremities of the object.
% , just as a 2D detector locates the object by its 2D extremities.
}
\label{fig:balance}
\end{figure}

\paragraph{Balancing Uncertainty and Discrimination} Consider the negative gradient \wrt the corresponding weights $w_i^\text{2D}$:
\begin{equation}
-\frac{\partial{L_\text{KL}}}{\partial{w_i^\text{2D}}} = w_i^\text{2D} \circ \left( -r_i^{\circ 2}(y_\text{gt}) + \hspace{-2ex} \expect_{y \sim p(y|X)}{\hspace{-1ex}r_i^{\circ 2}(y)} \right),
\end{equation}
where $r_i(y) = \pi(Rx_i^\text{3D} + t) - x_i^\text{2D}$ (unweighted reprojection error), and $(\cdot)^{\circ 2}$ stands for element-wise square. The first bracketed term $-r_i^{\circ 2}(y_\text{gt})$ with negative sign indicates that correspondences with large reprojection error (hence high uncertainty) shall be weighted less. The second term $\expect_{y \sim p(y|X)}{r_i^{\circ 2}(y)}$ is relevant to the variance of reprojection error over the predicted pose. The positive sign indicates that sensitive correspondences should be weighted more, because they provide stronger pose discrimination. The final gradient is thus a balance between the uncertainty and discrimination, as shown in Figure~\ref{fig:balance}. Existing work~\cite{monorun, pvnet} on learning uncertainty-aware correspondences only considers the former, hence lacking the discriminative ability.

\begin{algorithm}[t]
\DontPrintSemicolon
  \KwIn{$X = \{x_i^\text{3D}, x_i^\text{2D}, w_i^\text{2D}\}$}
  \KwOut{$L_\text{pred}$}
  $y^\ast, \Sigma_{y^\ast} \gets \mathit{PnP}(X)$ \tcp*{Laplace approximation}
  Fit $q_1(y)$ to $y^\ast, \Sigma_{y^\ast}$ \tcp*{initial proposal}
  
  \For{$1 \le t \le T$}{
    Generate $K^\prime$ samples $y^t_{j=1 \cdots K^\prime}$ from $q_t(y)$
    
    \For{$1 \le j \le K^\prime$}{
      $P_j^t \gets p(X|y_j^t)$ \tcp*{evaluate integrand}
    }
    
    \For{$1 \le \tau \le t$ \KwAnd $1 \le j \le K^\prime$}{
      $Q_j^\tau \gets \frac{1}{t}\sum_{m=1}^t q_m(y_j^\tau)$ \tcp*{eval proposal mix}
     
      $v_j^\tau \gets P_j^\tau / Q_j^\tau$  \tcp*{importance weight}
    }
    
    \If{$t < T$}{
    Estimate $q_{t+1}(y)$ from all weighted samples $\{y_j^\tau, v_j^\tau\ | \, 1 \le \tau \le t, 1 \le j \le K^\prime\}$}
  }
  $L_\text{pred} \gets \log \frac{1}{T K^\prime} \sum_{t=1}^{T} \sum_{j=1}^{K^\prime} v_j^t$

\caption{AMIS-based Monte Carlo pose loss}
\label{amisalg}
\end{algorithm}

\subsection{Local Regularization of Derivatives} \label{localreg}

While the KL divergence is a good metric for the probabilistic distribution, for inference it is still required to estimate the exact pose $y^\ast$ by solving the PnP problem in Eq.~(\ref{eqn:basicpnp}). The common choice of high precision is to utilize the iterative PnP solver based on the 
Levenberg-Marquardt (LM) algorithm -- a robust variant of the Gauss-Newton (GN) algorithm, which solves the non-linear least squares by the first and approximated second order derivatives. 
% However, as the learned pose distribution is potentially multimodal, LM iteration may be trapped at a local optimum. This motivates us to
To aid derivative-based optimization, we regularize the derivatives of the log density $\log p(y|X)$ \wrt the pose $y$, by encouraging the LM step $\Delta y$ to find the true pose $y_\text{gt}$.

To employ the regularization during training, a detached solution $y^\ast$ is obtained first. Then, at $y^\ast$, another iteration step is evaluated via the GN algorithm (which ideally equals 0 if $y^\ast$ has converged to the local optimum):
\begin{equation}
    \Delta y = -(J^\text{T}J + \varepsilon I)^{-1} J^\text{T} F(y^\ast),
    \label{gnstep}
\end{equation}
where $F(y^\ast) = \left[f_1^\text{T}(y^\ast), f_2^\text{T}(y^\ast), \cdots, f_N^\text{T}(y^\ast)\right]^\text{T}$ is the concatenated weighted reprojection errors of all points, $J = \partial{F(y)} / \partial{y^\text{T}} \negthickspace \bigm|_{y=y^\ast}$ is the Jacobian matrix, 
% $J^\text{T}J$ is an approximation to the Hessian matrix, 
and $\varepsilon$ is a small value for numerical stability. Note that $\Delta y$ is analytically differentiable. We therefore design the regularization loss as follows:
\begin{equation}
L_\text{reg} = l(y^\ast + \Delta y, y_\text{gt}),
\label{regloss}
\end{equation}
where $l(\cdot, \cdot)$ is a distance metric for pose. We adopt smooth L1 for position and cosine similarity for orientation (see supplementary materials for details). Note that the gradient is only backpropagated through $\Delta y$, encouraging the step to be non-zero if $y^\ast \neq y_\text{gt}$. 

This regularization loss can be also used as a standalone objective to train pose estimators~\cite{repose}. However, this objective alone cannot handle pose ambiguity properly, and is thus regarded as a secondary regularization in this paper.

\section{Attention-Inspired Correspondence Networks}

As discussed in Section~\ref{backprop}, the balance between uncertainty and discrimination enables locating important correspondences in an attention-like manner.
This inspires us to take elements from attention-related work, \ie, the Softmax layer and the deformable sampling~\cite{deformabledetr}.

In this section, we present two networks with EPro-PnP layer for 6DoF pose estimation and 3D object detection, respectively. For the former, EPro-PnP is incorporated into the existing dense correspondence architecture~\cite{CDPN}. For the latter, we propose a radical deformable correspondence network to explore the flexibility of EPro-PnP.

\subsection{Dense Correspondence Network} \label{densecorrnet}
For a strict comparison against existing PnP-based pose estimators, this paper takes the network from CDPN~\cite{CDPN} as a baseline, adding minor modifications to fit the EPro-PnP.

The original CDPN feeds cropped image regions within the detected 2D boxes into the pose estimation network, to which two decoupled heads are appended for rotation and translation respectively.
The rotation head is PnP-based while the translation head uses direct regression. 
This paper discards the translation head to focus entirely on PnP. 

\begin{figure}[t]
\begin{center}
    \includegraphics[width=0.9\linewidth]{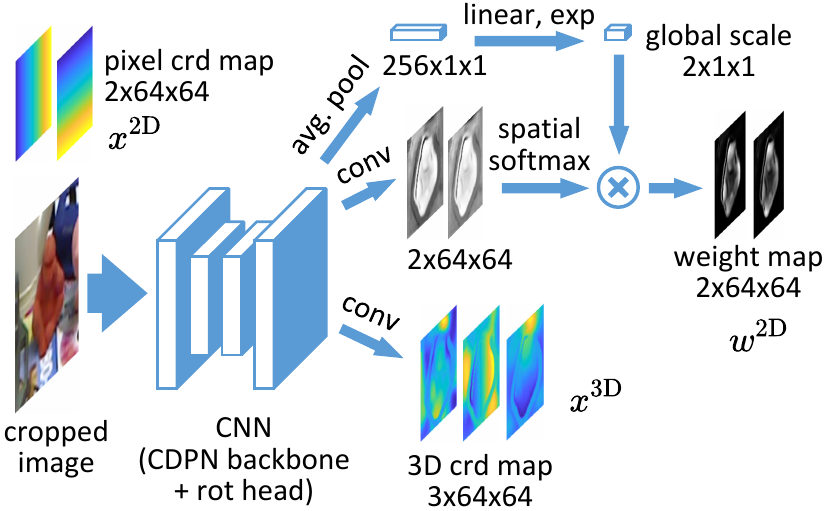}
\end{center}
\vspace{-1ex}
\caption{\textbf{The 6DoF pose estimation network} modified from CDPN~\cite{CDPN}. 
% The original single-channel confidence map is replaced by the two-channel weight map,
with spatial Softmax and global weight scaling.}
\label{fig:cdpn}
\end{figure}

Modifications are only made to the output layers. As shown in Figure~\ref{fig:cdpn}, the original confidence map is expanded to two-channel XY weights with spatial Softmax and dynamic global weight scaling. Inspired by the attention mechanism~\cite{vaswani2017attention}, the Softmax layer is a vital element for stable training, as it translates the absolute corresponding weights into a relative measurement. On the other hand, the global weight scaling factors represent the global concentration of the predicted pose distribution, ensuring a better convergence of the KL divergence loss.

The dense correspondence network can be trained solely with the KL divergence loss $L_\text{KL}$ to achieve decent performance. For top-tier performance, it is still beneficial to utilize additional coordinate regression as intermediate supervision, not to stabilize convergence but to introduce the geometric knowledge from the 3D models. Therefore, we keep the masked coordinate regression loss from CDPN~\cite{CDPN}
but leave out its confidence loss. Furthermore, the performance can be elevated by imposing the regularization loss $L_\text{reg}$ in Eq.~(\ref{regloss}).

\begin{figure*}[t]
   \begin{center}
   \includegraphics[width=0.98\textwidth]{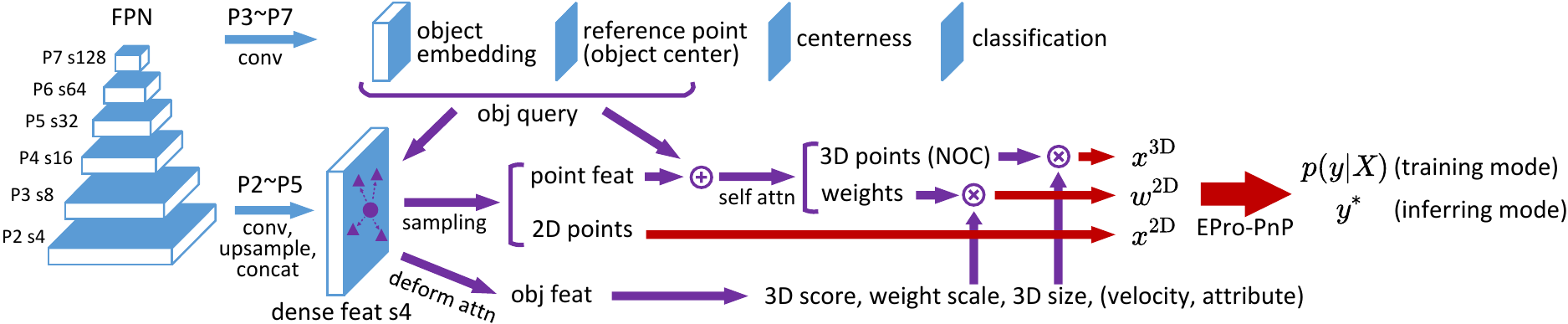}
   \end{center}
   \vspace{-2mm}
   \caption{\textbf{The deformable correspondence network} based on the FCOS3D~\cite{fcos3d} detector. Note that the sampled point-wise features are shared by the point-level subnet and the deformable attention layer that aggregates the features for object-level predictions.} 
\label{fig:fcos3d}
\end{figure*}

\subsection{Deformable Correspondence Network} \label{deformnetmain}
Inspired by Deformable DETR~\cite{deformabledetr}, we propose a novel deformable correspondence network for 3D object detection, in which the entire 2D-3D coordinates and weights are learned from scratch.

As shown in Figure~\ref{fig:fcos3d}, the deformable correspondence network is an extension of the FCOS3D~\cite{fcos3d} framework. The original FCOS3D is a one-stage detector that directly regresses the center offset, depth, and yaw orientation of multiple objects for 4DoF pose estimation. In our adaptation, the outputs of the multi-level FCOS head~\cite{fcos} are modified to generate object queries instead of directly predicting the pose. Also inspired by Deformable DETR~\cite{deformabledetr}, the appearance and position of a query is disentangled into the embedding vector and the reference point. A multi-head deformable attention layer~\cite{deformabledetr} is adopted to sample the key-value pairs from the dense features, with the value projected into \emph{point-wise features}, and meanwhile aggregated into the \emph{object-level features}. 

The point features are passed into a subnet that predicts the 3D points and corresponding weights (normalized by Softmax). Following MonoRUn~\cite{monorun}, the 3D points are set in the normalized object coordinate (NOC) space to handle categorical objects of various sizes.

The object features are 
responsible for predicting the object-level properties: (a) the 3D score (\ie, 3D localization confidence), (b) the weight scaling factor (same as in Section~\ref{densecorrnet}), (c) the 3D box size for recovering the absolute scale of the 3D points, and (d) other optional properties (velocity, attribute) required by the nuScenes benchmark~\cite{nuscenes}.

The deformable 2D-3D correspondences can be learned solely with the KL divergence loss $L_\text{KL}$,
% without any geometric prior,
preferably in conjunction with the regularization loss $L_\text{reg}$. Other auxiliary losses can be imposed onto the dense features for enhanced accuracy. Details are given in supplementary materials.

\section{Experiments}
\subsection{Datasets and Metrics}
\paragraph{LineMOD Dataset and Metrics}
The LineMOD dataset~\cite{linemod} consists of 13 sequences, each containing about 1.2K images annotated with 6DoF poses of a single object. Following \cite{Brachmann_2016_CVPR}, the images are split into the training and testing sets, with about 200 images per object for training. For data augmentation, we use the same synthetic data as in CDPN~\cite{CDPN}.
% , in which 1000 images per object are rendered with random background images from the PASCAL VOC2012 dataset.
We use two common metrics for evaluation: ADD(-S) and $n\text{\textdegree}, n\,\text{cm}$. The ADD measures whether the average deviation of the transformed model points is less than a certain fraction of the object’s diameter (\eg, ADD-0.1d). For symmetric objects, ADD-S computes the average distance to the closest model point. $n\text{\textdegree}, n\,\text{cm}$ measures the accuracy of pose based on angular/positional error thresholds. All metrics are presented as percentages.

% \paragraph{KITTI Dataset and Metrics}
% Thr KITTI dataset~\cite{kitti} consists of 7418 training images and 7518 testing images with multiple objects annotated with 3D bounding boxes. Following \cite{3dproposal}, the training set is further split by half into the training and validating subsets. As in \textbf{TBD}, only the detection performance on the \emph{Car} class is considered. Evaluation metrics are based on precision-recall curves with 3D IoU threshold of 0.7. The average precision (AP) is computed by the 40-point interpolation proposed in MonoDIS~\cite{monodis}.

\paragraph{nuScenes Dataset and Metrics}
The nuScenes 3D object detection benchmark~\cite{nuscenes} provides a large scale of data collected in 1000 scenes.
% , covering a variety of weather, time and location. 
Each scene contains 40 keyframes, annotated with a total of 1.4M 3D bounding boxes from 10 categories.
Each keyframe includes 6 RGB images collected from surrounding cameras. The data is split into 700/150/150 scenes for training/validation/testing. The official benchmark evaluates the average precision with true positives judged by 2D center error on the ground plane.
% instead of the 3D IoU. 
The mAP metric is computed by averaging over the thresholds of 0.5, 1, 2, 4 meters. Besides, there are 5 true positive metrics: Average Translation Error (ATE),
% , 2D center distance
Average Scale Error (ASE),
% , computed as $1-IoU$ with $IoU$ being the aligned 3D IoU
Average Orientation Error (AOE),
% , yaw angle difference in radians
Average Velocity Error (AVE)
% , L2-Norm of the error of the 2D velocity vector
and Average Attribute Error (AAE).
% , defined as $1 - acc$
Finally, there is a nuScenes detection score (NDS) computed as a weighted average of the above metrics. 

\subsection{Implementation Details} \label{implementation}
\paragraph{EPro-PnP Configuration} For the PnP formulation in Eq.~(\ref{eqn:basicpnp}), in practice the actual reprojection costs are robustified by the Huber kernel $\rho(\cdot)$:
\begin{equation}
\argmin_{y} \frac{1}{2} \sum_{i=1}^N \rho \left( \left\| f_i(y) \right\|^2 \right).
\label{eqn:robustpnp}
\end{equation}
The Huber kernel with threshold $\delta$ is defined as:
\begin{equation}
 \rho(s) = 
 \begin{dcases}
 s, & s \leq \delta^2,\\
 \delta(2 \sqrt{s} - \delta), & s > \delta^2.
 \end{dcases}
\label{eqn:huber}
\end{equation}
We use an adaptive threshold as described in the supplementary materials.
For Monte Carlo pose loss, we set the AMIS iteration count $T$ to 4 and the number of samples per iteration $K^\prime$ to 128. The loss weights are tuned such that $L_\text{KL}$ produces roughly the same magnitude of gradient as typical coordinate regression, while the gradient from $L_\text{reg}$ are kept very low. The weight normalization technique in \cite{monorun} is adopted to compute the dynamic loss weight for $L_\text{KL}$.
% We implement a pure PyTorch-based batch LM algorithm, which is faster than for-looping the Ceres-based CPU solver for a large batch.\footnote{The best efficiency could be achieved with an optimized CUDA implementation, which will be considered for the release version.}

\paragraph{Training the Dense Correspondence Network} 
General settings are kept the same as in CDPN~\cite{CDPN} (with ResNet-34~\cite{resnet} as backbone) for strict comparison, except that we increase the batch size to 32 for less training wall time.
The network is trained for 160 epochs by RMSprop on the LineMOD dataset~\cite{linemod}. To reduce the Monte Carlo overhead, 512 points are randomly sampled from the 64\texttimes64 dense points to compute $L_\text{KL}$.
% Initial learning rate is 0.0001, and stepped at epoch 50/100/150 with a decay factor of 0.1.

\paragraph{Training the Deformable Correspondence Network} We adopt the same detector architecture as in FCOS3D~\cite{fcos3d}, with ResNet-101-DCN~\cite{dcn} as backbone.
% , and disentangled FCOS heads. 
% Flip augmentation is also utilized during training. 
The network is trained for 12 epochs by the AdamW~\cite{adamw} optimizer, with a batch size of 12 images across 4 GPUs on the nuScenes dataset~\cite{nuscenes}.

\subsection{Results on the LineMOD Benchmark}

\paragraph{Comparison to the CDPN baseline with Ablations}
The contributions of every single modification to CDPN~\cite{CDPN} are revealed in Table~\ref{tab:baselinecomp}. 
% Before EPro-PnP is applied, we first remove the translation head, set the batch size to 32, and add the iterative PnP solver based on the LM algorithm. 
From the results it can be observed that:
\begin{itemize}[noitemsep,topsep=0.7ex,partopsep=0.7ex]
\item The original CDPN heavily relies on direct position regression, and the performance drops greatly (-17.46) when reduced to a pure PnP estimator, although the LM solver partially recovers the mean metric (+6.29). 
\item Employing EPro-PnP with the KL divergence loss significantly improves the metric (+13.84), outperforming CDPN-Full by a clear margin (65.88 \vs 63.21). 
\item The regularization loss proposed in Eq.~(\ref{regloss}) further elevates the performance (+1.88). 
\item Strong improvement (+5.46) is seen when initialized from A1, because CDPN has been trained with the extra ground truth of object masks, providing a good initial state highlighting the foreground.
\item Finally, the performance benefits (+0.97) from more training epochs (160 ep. from A1 + 320 ep.) as equivalent to CDPN-Full~\cite{CDPN} (3 stages \texttimes\ 160 ep.).
\end{itemize}
The results clearly demonstrate that EPro-PnP can unleash the enormous potential of the classical PnP approach, without any fancy network design or decoupling tricks. 

\paragraph{Comparison to the State of the Art}

As shown in Table~\ref{tab:sota}, despite modified from the lower baseline, EPro-PnP easily reaches comparable performance to the top pose refiner RePOSE~\cite{repose}, which adds extra overhead to the PnP-based initial estimator PVNet~\cite{pvnet}. Among all these entries, EPro-PnP is the most straightforward as it simply solves the PnP problem itself, without refinement network~\cite{repose, DPOD}, disentangled translation~\cite{CDPN,gdrnet}, or multiple representations~\cite{hybridpose}.

\paragraph{Comparison to Implicit Differentiation and Reprojection Learning} As shown in Table~\ref{tab:losscompare}, 
% the proposed Monte Carlo pose loss is compared against the implicit differentiation~\cite{BPnP} and the uncertainty-aware reprojection loss~\cite{monorun}.
% The experiments are conducted with only the loss functions substituted. 
% It can be observed that, 
when the coordinate regression loss is removed,
% \ie, object 3D models are unavailable, 
both implicit differentiation and reprojection loss fail to learn the pose properly. Yet EPro-PnP manages to learn the coordinates from scratch, even outperforming CDPN without translation head (79.46 \vs 74.54). 
% Only with coordinate regression involved can the other methods produce reasonable results. 
This validates that EPro-PnP can be used as a general pose estimator without relying on geometric prior.

\paragraph{Uncertainty and Discrimination}
In Table~\ref{tab:losscompare}, \textit{Reprojection} \vs \textit{Monte Carlo} loss can be interpreted as uncertainty alone \vs uncertainty-discrimination balanced. The results reveal that uncertainty alone exhibits strong performance when intermediate coordinate supervision is available, while discrimination is the key element for learning correspondences from scratch.

\paragraph{Contribution of End-to-End Weight/Coordinate Learning}

As shown in Table~\ref{tab:baselinecomp}, 
% we evaluate the component-wise contribution of the end-to-end pose loss by detaching the weights or coordinates from the loss function. It can be observed that
detaching the weights from the end-to-end loss has a stronger impact to the performance than detaching the coordinates (\textminus8.69 \vs \textminus3.08), stressing the importance of attention-like end-to-end weight learning.

\paragraph{On the Importance of the Softmax Layer} Learning the corresponding weights without the normalization denominator of spatial Softmax (so it becomes exponential activation as in \cite{monorun}) does not converge, as listed in Table~\ref{tab:baselinecomp}.

\begin{table}[t]
    \RawFloats
    % \vspace*{-2mm}
    \begin{center}
    \scalebox{0.8}{%
    \setlength{\tabcolsep}{0.4em}
    \begin{tabular}{clcccl}
        \toprule
        \multirow{2}[2]{*}{ID} &         \multirow{2}[2]{*}{Method} & \multicolumn{3}{c}{ADD(-S)} & \multirow{2}[2]{*}{Mean} \\
        \cmidrule(lr){3-5}
        {} & {} & 0.02d & 0.05d & 0.1d \\
        \midrule
        A0 & CDPN-Full~\cite{CDPN} & 29.10 & 69.50 & 91.03 & 63.21 \\
        A1 & CDPN w/o trans. head & 15.93 & 46.79 & 74.54 & 45.75 (\textminus17.46) \\
        A2 & + Batch=32, LM solver & 21.17 & 55.00 & 79.96 & 52.04 (+\phantom{0}6.29) \\
        \midrule
        B0 & Basic EPro-PnP & 32.14 & 72.83 & 92.66 & 65.88 (+13.84) \\
        B1 & +\ Regularize derivatives & 35.44 & 74.41 & 93.43 & 67.76 (+\phantom{0}1.88) \\
        B2 & +\ Initialize from A1 & 42.92 & 80.98 & 95.76 & 73.22 (+\phantom{0}5.46) \\
        B3 & +\ Long sched. (320 ep.) & 44.81 & 81.96 & 95.80 & 74.19 (+\phantom{0}0.97) \\
        \midrule
        C0 & B0 \textrightarrow\ Detach coords. & 29.57 & 68.61 & 90.23 & 62.80 (\textminus\phantom{0}3.08) \\
        C1 & B0 \textrightarrow\ Detach weights & 22.99 & 61.31 & 87.27 & 57.19 (\textminus\phantom{0}8.69) \\
        \midrule
        D0 & B0 \textrightarrow\ No Softmax denom.\hspace{-2ex} & \multicolumn{4}{c}{divergence} \\
        \bottomrule
    \end{tabular}}
    \end{center}
    \vspace{-1ex}
    \caption{\textbf{Comparison to the CDPN baseline with Ablation Studies.} Results of CDPN are reproduced with the official code.\protect\footnotemark\ In C0/C1 either component is detached individually from the KL loss, while adding a surrogate mask regression loss~\cite{CDPN} in C1.} 
    \label{tab:baselinecomp}
    
    \begin{center}
    \scalebox{0.8}{%
    \setlength{\tabcolsep}{0.5em}
    \begin{tabular}{lccccc}
        \toprule
        \multirow{2}[2]{*}{Method} & \multirow{2}[2]{*}{2\textdegree, 2 cm} & \multirow{2}[2]{*}{5\textdegree, 5 cm} & \multicolumn{3}{c}{ADD(-S)} \\
        \cmidrule(lr){4-6}
        {} & {} & {} & 0.02d & 0.05d & 0.1d \\
        \midrule
        CDPN~\cite{CDPN} & - & 94.31 & - & - & 89.86 \\
        HybridPose~\cite{hybridpose} & - & - & - & - & 91.3\phantom{0} \\
        % BPnP~\cite{BPnP} & - & - & - & - & \textbf{93.27} \\
        GDRNet*~\cite{gdrnet} & 67.1\phantom{0} & - & 35.6\phantom{0} & 76.0\phantom{0} & 93.6\phantom{0} \\
        DPOD~\cite{DPOD} & - & - & - & - & 95.15 \\
        PVNet-RePOSE~\cite{repose} & - & - & - & - & 96.1\phantom{0} \\
        \midrule
        EPro-PnP & 80.99 & 98.54 & 44.81 & 81.96 & 95.80 \\
        \bottomrule
    \end{tabular}}
    \end{center}
    \vspace{-1ex}
    \caption{\textbf{Comparison to the state-of-the-art geometric methods.} BPnP~\cite{BPnP} is not included as it adopts a different train/test split. *Although GDRNet~\cite{gdrnet} only reports the performance in its ablation section, it is still a fair comparison to our method, since both use the same baseline (CDPN).
    }
    \label{tab:sota}
    
    \begin{center}
    \scalebox{0.8}{%
    \setlength{\tabcolsep}{0.5em}
    \begin{tabular}{lcccccc}
        \toprule
        Main Loss & \makecell{Coord. \\ Regr.} & 2\textdegree & 2 cm & 2\textdegree, 2 cm & \makecell{ADD(-S) \\ 0.1d} \\
        \midrule
        Implicit diff.~\cite{BPnP} & {} & \multicolumn{4}{c}{divergence} \\
        Reprojection~\cite{monorun} & {} & 0.32 & 42.30 & 0.16 & 14.56 \\
        Monte Carlo (ours) & {} & 44.18 & 81.55 & 40.96 & 79.46 \\
        \midrule
        Implicit diff.~\cite{BPnP} & \checkmark & 56.13 & 91.13 & 53.33 & 88.74 \\
        Reprojection~\cite{monorun} & \checkmark & 62.79 & 92.91 & 60.65 & 92.04 \\
        Monte Carlo (ours) & \checkmark & 65.75 & 93.90 & 63.80 & 92.66 \\
        \bottomrule
    \end{tabular}}
    \end{center}
    \vspace{-1ex}
    \caption{\textbf{Comparison between loss functions} by experiments conducted on the same dense correspondence network.
    % without regularization loss, pretrained weights or longer epochs
    For implicit differentiation, we minimize the distance metric of pose in Eq.~(\ref{regloss}) instead of the reprojection-metric pose loss in BPnP~\cite{BPnP}.}
    \label{tab:losscompare}
\end{table}

\begin{figure*}[t]
\begin{floatrow}
\CenterFloatBoxes
\capbtabbox{%
    \scalebox{0.8}{%
    \setlength{\tabcolsep}{0.5em}
    \begin{tabular}{lcccccccc}
        \toprule
        \multirow{2}[2]{*}{Method} & \multirow{2}[2]{*}{Data} & \multirow{2}[2]{*}{NDS} & \multirow{2}[2]{*}{mAP} & \multicolumn{5}{c}{True positive metrics (lower is better)} \\
        \cmidrule(lr){5-9}
        {} & {} & {} & {} & mATE & mASE & mAOE & mAVE & mAAE \\
        \midrule
        CenterNet~\cite{centernet} & Val & 0.328 & 0.306 & 0.716 & 0.264 & 0.609 & 1.426 & 0.658 \\
        FCOS3D\cite{fcos3d} & Val & 0.372 & 0.295 & 0.806 & 0.268 & 0.511 & 1.315 & 0.170 \\
        % DETR3D~\cite{detr3d} & Val & 0.374 & 0.303 & 0.860 & 0.278 & 0.437 & 0.967 & 0.235 \\
        % FCOS3D\S*~\cite{fcos3d} & Val & 0.402 & 0.326 & 0.743 & 0.259 & 0.441 & 1.341 & 0.163 \\
        FCOS3D\S\textdagger~\cite{fcos3d} & Val & 0.415 & 0.343 & 0.725 & 0.263 & 0.422 & 1.292 & \textbf{0.153} \\
        PGD\S~\cite{pgd} & Val & 0.422 & \textbf{0.361} & 0.694 & 0.265 & 0.442 & 1.255 & 0.185 \\
        \midrule
        Basic EPro-PnP & Val & 0.425 & 0.349 & 0.676 & 0.263 & 0.363 & 1.035 & 0.196 \\
        + coord. regr. & Val & 0.430 & 0.352 & 0.667 & 0.258 & 0.337 & 1.031 & 0.193 \\
        + TTA\S & Val & \textbf{0.439} & \textbf{0.361} & \textbf{0.653} & \textbf{0.255} & \textbf{0.319} & \textbf{1.008} & 0.193 \\
        \midrule
        MonoDIS~\cite{monodis} & Test & 0.384 & 0.304 & 0.738 & 0.263 & 0.546 & 1.553 & 0.134 \\
        CenterNet~\cite{centernet} & Test & 0.400 & 0.338 & 0.658 & 0.255 & 0.629 & 1.629 & 0.142 \\
        FCOS3D\S\textdagger~\cite{fcos3d} & Test & 0.428 & 0.358 & 0.690 & 0.249 & 0.452 & 1.434 & \textbf{0.124} \\
        PGD\S~\cite{pgd} & Test & 0.448 & \textbf{0.386} & 0.626 & 0.245 & 0.451 & 1.509 & 0.127 \\
        \midrule
        EPro-PnP\S & Test & \textbf{0.453} & 0.373 & \textbf{0.605} & \textbf{0.243} & \textbf{0.359} & \textbf{1.067} & \textbf{0.124} \\
        \bottomrule
    \end{tabular}}
    }{%
    \vspace{2mm}
    \caption{\textbf{3D object detection results} on the nuScenes benchmark. Methods with extra pretraining other than ImageNet backbone are not included for comparison. \S\ indicates test-time flip augmentation (TTA). \textdagger\ indicates model ensemble. 
    % *Note that the difference between FCOS3D and FCOS3D\S\ is more than just TTA, see \cite{fcos3d} for details.
    }
    \label{tab:nuscenes}}
% \hspace*{\fill}
\hspace*{-3ex}
\ffigbox[0.76\FBwidth][][t]{%
  \includegraphics[width=1.0\linewidth]{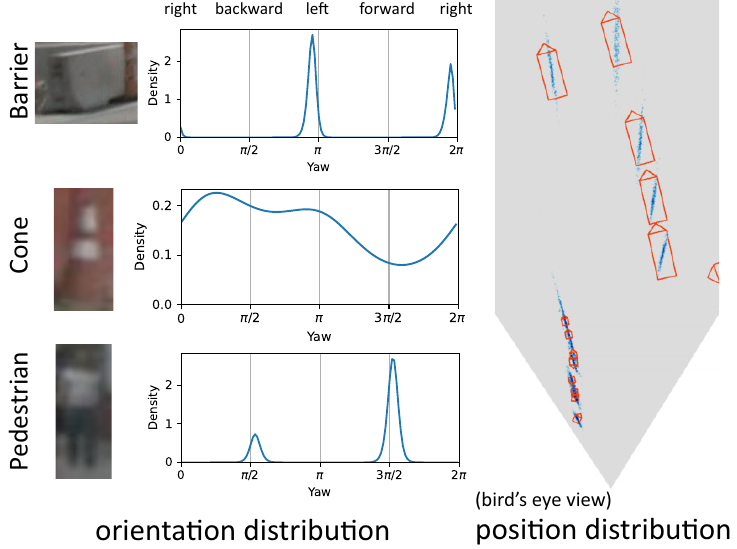}
}{%
  \vspace{-2.6mm}
  \caption{
  \textbf{Visualization of the predicted pose distribution}. The orientation density is clearly multimodal, capturing the pose ambiguity of symmetric objects (\textit{Barrier}, \textit{Cone}) and uncertain observations (\textit{Pedestrian}).}
  \label{fig:nus_viz}
}
\end{floatrow}
\vspace*{-0.5ex}
\end{figure*}

\subsection{Results on the nuScenes Benchmark}

We evaluate 3 variants of EPro-PnP: (a) the basic approach that learns deformable correspondences without geometric prior (enhanced with regularization), (b) adding coordinate regression loss with sparse ground truth extracted from the available LiDAR points as in \cite{monorun}, (c) further adding test-time flip augmentation (TTA) for fair comparison against \cite{fcos3d, pgd}. All results on the validation/test sets are presented in Table~\ref{tab:nuscenes} with comparison to other approaches.

\addtocounter{footnote}{-1}
 \stepcounter{footnote}\footnotetext{\url{https://git.io/JXZv6}}
%  \stepcounter{footnote}\footnotetext{}

% Table~\ref{tab:nuscenes}
% shows the 3D detection performance of the proposed deformable correspondence network trained with EPro-PnP, as well as other entries from the official leaderboard.
From the validation results it can be observed that:
\begin{itemize}[noitemsep,topsep=0.7ex,partopsep=0.7ex]
\item The basic EPro-PnP significantly outperforms the FCOS3D~\cite{fcos3d} baseline (NDS 0.425 \vs 0.372). Although it partially benefits from more parameters from the correspondence head, there is still good evidence that: with a proper end-to-end pipeline, PnP can outperform direct pose prediction on a large scale of data.
\item Regarding the mATE and mAOE metrics that reflect pose accuracy, the basic EPro-PnP already outperforms all previous methods, again demonstrating that EPro-PnP is a better pose estimator. The coordinate regression loss helps further reducing the orientation error (mAOE 0.337 \vs 0.363).
\item With TTA, EPro-PnP outperforms the state of the art by a clear margin (NDS 0.439 \vs 0.422) on the validation set.
\end{itemize}

On the test data, with the advantage in pose accuracy (mATE and mAOE), EPro-PnP achieves the highest NDS score among other task-specific competitors.

\subsection{Qualitative Analysis}

\begin{figure}[t]
\vspace*{-3.5ex}
\begin{center}
\hspace{-4mm}\includegraphics[width=0.93\linewidth]{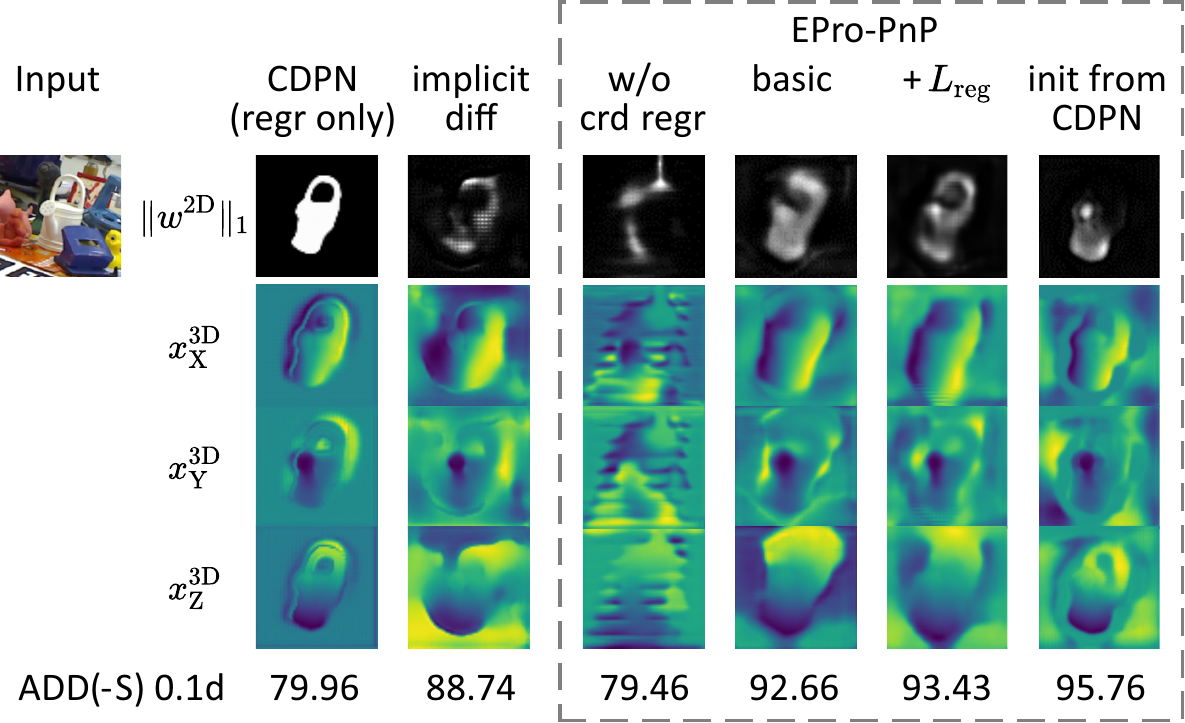}
\end{center}
\vspace*{-2mm}
\caption{\textbf{Visualization of the inferred weight and coordinate maps} on LineMOD test data.} 
\label{fig:cdpn_viz}
\end{figure}

As illustrated in Figure~\ref{fig:cdpn_viz}, the dense weight and coordinate maps learned with EPro-PnP generally capture less details compared to CDPN~\cite{CDPN}, as a result of higher uncertainty around sharp edges. Surprisingly, even though the learned-from-scratch coordinate maps seem to be a mess, the end-to-end pipeline gains comparable pose accuracy to the CDPN baseline (79.46 \vs 79.96). When initialized with pretrained CDPN, EPro-PnP inherits the detailed geometric profile, therefore confining the active weights within the foreground region and achieving the overall best performance. Also note that the weight maps of both derivative regularization and implicit differentiation~\cite{BPnP} are more concentrated, biasing towards discrimination over uncertainty.

Figure~\ref{fig:nus_viz} shows that the flexibility of EPro-PnP allows predicting multimodal distributions with strong expressive power, successfully capturing the orientation ambiguity without discrete multi-bin classification~\cite{fcos3d,mousavian20173d} or complicated mixture model~\cite{bui20206d}. Owing to the ability to model orientation ambiguity, EPro-PnP outperforms other competitors by a wide margin in terms of the AOE metric in Table~\ref{tab:nuscenes}.

\section{Conclusion}

This paper proposes the EPro-PnP, which translates the non-differentiable deterministic PnP operation into a differentiable probabilistic layer, empowering end-to-end 2D-3D correspondence learning of unprecedented flexibility. The connections to previous work~\cite{monorun,BPnP,blindpnp,dsac++} have been thoroughly discussed with theoretical and experimental proofs. For application, EPro-PnP can inspire novel solutions such as the deformable correspondence, or it can be simply integrated into existing PnP-based networks.
% , reaching top-tier accuracy without tricky task-specific design. 
Beyond the PnP problem, the underlying principles are theoretically generalizable to other learning models with nested optimization layer, known as declarative networks~\cite{declarative}.

\paragraph{Acknowledgements} This research was supported by Alibaba Group through Alibaba Research Intern Program, the National Natural Science Foundation of China (No.~52002285), the Shanghai Pujiang Program (No.~2020PJD075), the Natural Science Foundation of Shanghai (No.~21ZR1467400), and the Perspective Study Funding of Nanchang Automotive Institute of Intelligence \& New Energy, Tongji University (TPD-TC202110-03).

%%%%%%%%% REFERENCES
{\small
\bibliographystyle{ieee_fullname}
\bibliography{egbib}
}

%%%%%%%%% supplementary
\clearpage
\setcounter{section}{0}
\renewcommand{\thesection}{\Alph{section}}

\section{Levenberg-Marquardt PnP Solver}

For scalability, we have implemented a PyTorch-based batch Levenberg-Marquardt (LM) PnP solver. The implementation generally follows the Ceres solver~\cite{ceres-solver}. Here, we discuss some important details that are related to the proposed Monte Carlo pose sampling and derivative regularization.

\subsection{Adaptive Huber Kernel}

To robustify the weighted reprojection errors of various scales, we adopt an adaptive Huber kernel with a dynamic threshold $\delta$ for each object, defined as a function of the weights ${w^\text{2D}_i}$ and 2D coordinates ${x^\text{2D}_i}$:
\begin{equation}
    \delta = \delta_\text{rel} \frac{\left\| \bar{w}^\text{2D} \right\|_1}{2}  \left( \frac{1}{N-1}\sum_{i=1}^N{\left\|x^\text{2D}_i - \bar{x}^\text{2D}\right\|^2} \right)^{\negthickspace\frac{1}{2}},
\end{equation}
with the relative threshold $\delta_\text{rel}$ as hyperparameter, and the mean vectors $\bar{w}^\text{2D}=\frac{1}{N}\sum_{i=1}^N{w_i^\text{2D}},\, \bar{x}^\text{2D}=\frac{1}{N}\sum_{i=1}^N{x_i^\text{2D}}$.

\subsection{LM Step with Huber Kernel}

Adding the Huber kernel influences every related element from the likelihood function to the LM iteration step and derivative regularization loss. Thanks to PyTorch's automatic differentiation, the robustified Monte Carlo KL divergence loss does not require much special handling. For the LM solver, however, the residual $F(y)$ (concatenated weighted reprojection errors) and the Jacobian matrix $J$ have to be rescaled before computing the robustified LM step~\cite{triggsba}.

The rescaled residual block $\tilde{f}_i(y)$ and Jacobian block $\tilde{J}_i(y)$ of the $i$-th point pair are defined as:
\begin{equation}
    \tilde{f}_i(y) = \sqrt{\rho^\prime_i} f_i(y),
\end{equation}
\begin{equation}
    \tilde{J}_i(y) = \sqrt{\rho^\prime_i} J_i(y),
\end{equation}
where
\begin{equation}
 \rho^\prime_i = 
 \begin{dcases}
 1, & \| f_i(y) \| \leq \delta,\\
 \frac{\delta}{\| f_i(y) \|}, & \| f_i(y) \| > \delta,
 \end{dcases}
\label{eqn:huber2}
\end{equation}
\begin{equation}
    J_i(y) = \frac{\partial{f_i(y)}}{\partial{y^\text{T}}}.
\end{equation}
Following the implementation of Ceres solver~\cite{ceres-solver}, the robustified LM iteration step is:
\begin{equation}
    \Delta y = -\left(\tilde{J}^\text{T}\tilde{J} + \lambda D^2 \right)^{-1} \tilde{J}^\text{T} \tilde{F},
    \label{eqn:robustlmstep}
\end{equation}
where
\begin{equation}
    \tilde{J} = 
    \begin{bmatrix}
        \tilde{J}_1(y) \\
        \vdots\\
        \tilde{J}_N(y)
    \end{bmatrix},
        \tilde{F} = 
    \begin{bmatrix}
        \tilde{f}_1(y) \\
        \vdots\\
        \tilde{f}_N(y)
    \end{bmatrix},
\end{equation}
$D$ is the square root of the diagonal of the matrix $\tilde{J}^\text{T}\tilde{J}$, and $\lambda$ is the reciprocal of the LM trust region radius~\cite{ceres-solver}.
% , and $H^\text{LM}$ compactly denotes the LM Hessian Matrix.

Note that the rescaled residual and Jacobian affects the derivative regularization (Eq.~(\ref{regloss})), as well as the covariance estimation in the next subsection.

\paragraph{Fast Inference Mode} We empirically found that in a well-trained model, the LM trust region radius can be initialized with a very large value, effectively rendering the LM algorithm redundant. We therefore use the simple Gauss-Newton implementation for fast inference:
\begin{equation}
    \Delta y = -\left(\tilde{J}^\text{T}\tilde{J} + \varepsilon I \right)^{-1} \tilde{J}^\text{T} \tilde{F},
\end{equation}
where $\varepsilon$ is a small value for numerical stability.

\subsection{Covariance Estimation}
During training, the concentration of the AMIS proposal is determined by the local estimation of pose covariance matrix $\Sigma_{y^\ast}$, defined as:
\begin{equation}
    \Sigma_{y^\ast} = \left( \tilde{J}^\text{T} \tilde{J} + \varepsilon I \right)^{-1} \Big\rvert_{y=y^\ast},
\end{equation}
where $y^\ast$ is the LM solution that determines the location of the proposal distribution.

\subsection{Initialization}
\label{rslm}

Since the LM solver only finds a local solution, initialization plays a determinant role in dealing with ambiguity. Standard EPnP~\cite{EPnP} initialization can handle the dense correspondence network trained on the LineMOD~\cite{linemod} dataset, where ambiguity is not noticeable. For the deformable correspondence network trained on the nuScenes~\cite{nuscenes} dataset and more general cases, we implement a random sampling algorithm analogous to RANSAC, to search for the global optimum efficiently.

Given the $N$-point correspondence set $X = \left\{x^\text{3D}_i,x^\text{2D}_i,w^\text{2D}_i\,\middle|\,i=1\cdots N\right\}$, we generate $M$ subsets consisting of $n$ corresponding points each ($3 \leq n < N$), by repeatedly sub-sampling $n$ indices without replacement from a multinomial distribution, whose probability mass function $p(i)$ is defined by the corresponding weights:
\begin{equation}
    p(i) = \frac{\left\|w^\text{2D}_i\right\|_1}{\sum_{i=1}^N{\left\|w^\text{2D}_i\right\|_1}}.
\end{equation}
From each subset, a pose hypothesis can be solved via the LM algorithm with very few iterations (we use 3 iterations). This is implemented as a batch operation on GPU, and is rather efficient for small subsets. We take the hypothesis of maximum log-likelihood $\log{p(X|y)}$ as the initial point, starting from which subsequent LM iterations are computed on the full set $X$. 

\paragraph{Training Mode}
During training, the LM PnP solver is utilized for estimating the location and concentration of the initial proposal distribution in the AMIS algorithm. The location is very important to the stability of Monte Carlo training. If the LM solver fails to find the global optimum and the location of the local optimum is far from the true pose $y_\text{gt}$, the balance between the two opposite signed terms in Eq.~(\ref{symbloss}) may be broken, leading to exploding gradient in the worst case scenario. To avoid such problem, we adopt a simple initialization trick: we compare the log-likelihood $\log{p(X|y)}$ of the ground truth $y_\text{gt}$ and the selected hypothesis, and then keep the one with higher likelihood as the initial state of the LM solver. 

\section{Details on Monte Carlo Pose Sampling}

\subsection{Proposal Distribution for Position}

For the proposal distribution of the translation vector $t \in \mathbb{R}^3$, we adopt the multivariate t-distribution, with the following probability density function (PDF):
\begin{equation}
    q_\text{T}(t) = \frac{\Gamma\left(\frac{\nu + 3}{2}\right)}{\Gamma\left(\frac{\nu}{2}\right) \sqrt{\nu^3 \pi^3 |\Sigma|}} \left( 1 + \frac{1}{\nu} \| t-\mu \|_{\Sigma}^2 \right)^{\negmedspace -\frac{\nu+3}{2}},
\end{equation}
where $\| t-\mu \|_{\Sigma}^2 = (t-\mu)^\text{T} \Sigma^{-1} (t-\mu)$, with the location $\mu$, the 3\texttimes3 positive definite scale matrix $\Sigma$, and the degrees of freedom $\nu$. Following~\cite{amis}, we set $\nu$ to 3. Compared to the multivariate normal distribution, the t-distribution has a heavier tail, which is ideal for robust sampling. 

The multivariate t-distribution has been implemented in the Pyro~\cite{pyro} package.

\paragraph{Initial Parameters}
The initial location and scale is determined by the PnP solution and covariance matrix, \ie, $\mu \gets t^\ast, \Sigma \gets \Sigma_{t^\ast}$, where $\Sigma_{t^\ast}$ is the 3\texttimes 3 submatrix of the full pose covariance $\Sigma_{p^\ast}$. Note that the actual covariance of the t-distribution is thus $\frac{\nu}{\nu-1}\Sigma_{t^\ast}$, which is intentionally scaled up for robust sampling in a wider range. 

\paragraph{Parameter Estimation from Weighted Samples}
To update the proposal, we let the location $\mu$ and scale $\Sigma$ be the first and second moment of the weighted samples (\ie, weighted mean and covariance), respectively.

\subsection{Proposal Distribution for 1D Orientation}
For the proposal distribution of the 1D yaw-only orientation $\theta$, we adopt a mixture of von Mises and uniform distribution. The von Mises is also known as the circular normal distribution, and its PDF is given by:
\begin{equation}
    q_\text{VM}(\theta) = \frac{\exp{(\kappa \cos{(\theta - \mu)})}}{2 \pi I_0(\kappa)},
\end{equation}
where $\mu$ is the location parameter, $\kappa$ is the concentration parameter, and $I_0(\cdot)$ is the modified Bessel function with order zero. The mixture PDF is thus:
\begin{equation}
    q_\text{mix}(\theta) = (1-\alpha)q_\text{VM}(\theta) + \alpha q_\text{uniform}(\theta),
\end{equation}
with the uniform mixture weight $\alpha$. The uniform component is added in order to capture other potential modes under orientation ambiguity. We set $\alpha$ to a fixed value of $1/4$.

PyTorch has already implemented the von Mises distribution, but its random sample generation is rather slow. As an alternative we use the NumPy implementation for random sampling.

\paragraph{Initial Parameters}
With the yaw angle $\theta^\ast$ and its variance $\sigma^2_{\theta^\ast}$ from the PnP solver, the parameters of the von Mises proposal is initialized by $\mu \gets \theta^\ast, \kappa \gets \frac{1}{3\sigma^2_{\theta^\ast}}$.

\paragraph{Parameter Estimation from Weighted Samples}
For the location $\mu$, we simply adopt its maximum likelihood estimation, \ie, the circular mean of the weighted samples. For the concentration $\kappa$, we first compute an approximated estimation~\cite{Dhillon2003ModelingDU} by:
\begin{equation}
    \hat{\kappa} = \frac{\bar{r}(2-\bar{r}^2)}{1-\bar{r}^2},
\end{equation}
where $\bar{r} = \left\lVert \sum_j v_j[\sin{\theta_j}, \cos{\theta_j}]^\text{T} / \sum_j v_j \right\rVert$ is the norm of the mean orientation vector, with the importance weight $v_j$ for the $j$-th sample $\theta_j$. Finally, the concentration is scaled down for robust sampling, such that $\kappa \gets \hat{\kappa}/3$.

\subsection{Proposal Distribution for 3D Orientation} Regarding the quaternion based parameterization of 3D orientation, which can be represented by a unit 4D vector $l$, we adopt the angular central Gaussian (ACG) distribution as the proposal. The support of the 4-dimensional ACG distribution is the unit hypersphere, and the PDF is given by:
\begin{equation}
    q_\text{ACG}(l) = \frac{(l^\text{T} \Lambda^{-1} l)^{-2}}{S_4 |\Lambda|^{\frac{1}{2}}},
\end{equation}
where $S_4 = 2 \pi^2$ is the 3D surface area of the 4D sphere, and $\Lambda$ is a 4\texttimes 4 positive definite matrix.

The ACG density can be derived by integrating the zero-mean multivariate normal distribution $\mathcal{N}(0, \Lambda)$ along the radial direction from $0$ to $\inf$. Therefore, drawing samples from the ACG distribution is equivalent to sampling from $\mathcal{N}(0, \Lambda)$ and then normalizing the samples to unit radius.

\paragraph{Initial Parameters}
Consider $l^\ast$ to be the PnP solution and $\Sigma_{l^\ast}^{-1}$ to be the estimated 4\texttimes 4 inverse covariance matrix. Note that $\Sigma_{l^\ast}^{-1}$ is only valid in the local tangent space with rank 3, satisfying ${l^\ast}^\text{T} \Sigma_{l^\ast}^{-1} l^\ast = 0$. The initial parameters are determined by:
\begin{equation}
    \Lambda \gets \hat{\Lambda} + \alpha |\hat{\Lambda}|^{\frac{1}{4}}I,
    \label{acgparam}
\end{equation}
where $\hat{\Lambda} = \left(\Sigma_{l^\ast}^{-1} + I\right)^{-1}$, and $\alpha$ is a hyperparameter that controls the dispersion of the proposal for robust sampling. We set $\alpha$ to 0.001 in the experiments.

\paragraph{Parameter Estimation from Weighted Samples}
Based on the samples $l_j$ and weights $v_j$, the maximum likelihood estimation $\hat{\Lambda}$ is the solution to the following equation:
\begin{equation}
    \hat{\Lambda} = \frac{4}{\sum_j v_j} \sum_j \frac{v_j l_j l_j^\text{T}}{l_j^\text{T} \hat{\Lambda}^{-1} l_j}.
    \label{acgmle}
\end{equation}
The solution to Eq.~(\ref{acgmle}) can be computed by fixed-point iteration~\cite{ACG}. The final parameters of the updated proposal is determined the same way as in Eq.~(\ref{acgparam}).

\section{Details on Derivative Regularization Loss}

As stated in the main paper, the derivative regularization loss $L_\text{reg}$ consists of the position loss $L_\text{pos}$ and the orientation loss $L_\text{orient}$. 

For $L_\text{pos}$, we adopt the smooth L1 loss based on the Euclidean distance $d_t = \| t^\ast + \Delta t - t_\text{gt} \|$, given by:
\begin{equation}
    L_\text{pos} = 
    \begin{dcases}
    \frac{d_t^2}{2\beta}, & d_t \leq \beta,\\
    d_t - 0.5\beta, & d_t > \beta,
    \end{dcases}
\end{equation}
with the hyperparameter $\beta$.

For $L_\text{orient}$, we adopt the cosine similarity loss based on the angular distance $d_\theta$. For 1D orientation parameterized by the angle $\theta$, $d_\theta = \theta^\ast + \Delta \theta - \theta_\text{gt}$. For 3D orientation parameterized by the quaternion vector $l$, $d_\theta = 2 \arccos{(l^\ast + \Delta l)^\text{T} l_\text{gt}}$. The loss function is therefore defined as:
\begin{equation}
    L_\text{orient} = 1 - \cos{d_\theta}.
    \label{orientloss}
\end{equation}
For 3D orientation, after the substitution, the loss function can be simplified to:
\begin{equation}
    L_\text{orient} = 2 - 2 \left( (l^\ast + \Delta l)^\text{T} l_\text{gt} \right)^2.
\end{equation}

For the specific settings of the hyperparameter $\beta$ and loss weights, please refer to the experiment configuration code.

\section{Details on the Deformable Correspondence Network}

\subsection{Network Architecture} \label{deformnetsup}
\begin{figure}[t]
\begin{center}
    \includegraphics[width=0.95\linewidth]{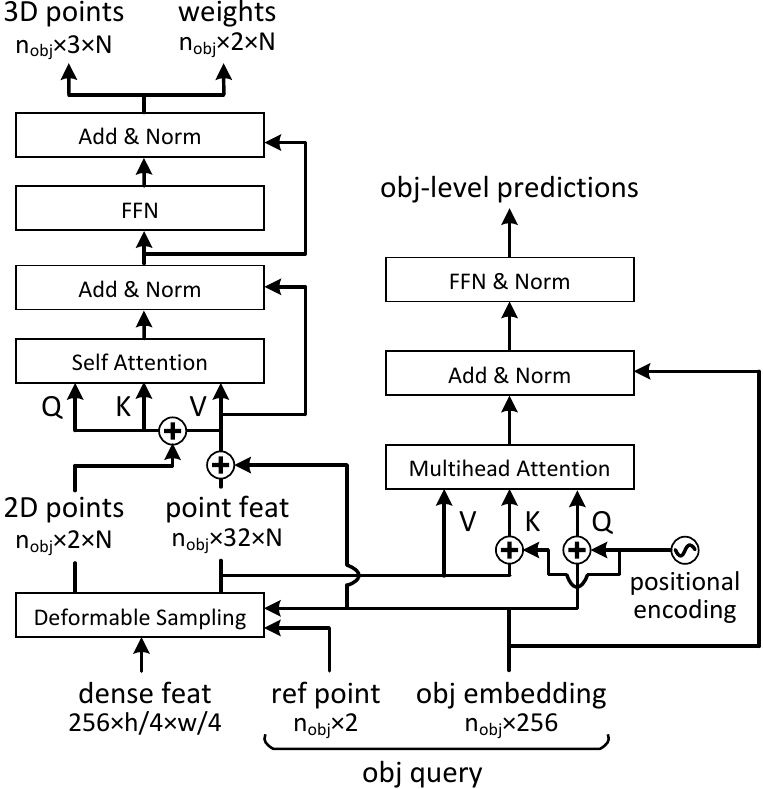}
\end{center}
\vspace{-1ex}
\caption{Detailed architecture of the deformable correspondence network.}
\label{fig:deformcorr}
\end{figure}

The detailed network architecture of the deformable correspondence network is shown in Figure~\ref{fig:deformcorr}.
% The design is largely inspired by the DETR family~\cite{detr,deformabledetr} of transformer-based object detectors. 
% In fact, the deformable correspondence network could have been integrated into the DETR framework. However, there was no DETR-like baseline for monocular 3D object detection before the very recent DETR3D~\cite{detr3d}, so we decide on the FCOS3D~\cite{fcos3d} framework as an alternative. 
Following deformable DETR~\cite{deformabledetr}, this paper adopts the multi-head deformable sampling. Let $n_\text{head}$ be the number of heads and $n_\text{hpts}$ be the number of points per head, a total number of $N = n_\text{head} n_\text{hpts}$ points are sampled for each object. The sampling locations relative to the reference point are generated from the object embedding by a single layer of linear transformation. We set $n_\text{head}$ to 8, which yields $256 / n_\text{head} = 32$ channels for the point features.

The point-level branch on the left side of Figure~\ref{fig:deformcorr} is responsible for predicting the 3D points $x^\text{3D}_i$ and corresponding weights $w^\text{2D}_i$. The sampled point features are first enhanced by the object-level context, by adding the reshaped head-wise object embedding to the point features. Then, the features of the $N$ points are processed by the self attention layer, for which the 2D points are transformed into positional encoding. The attention layer is followed by standard layers of normalization, skip connection, and feedforward network (FFN).

Regarding the object-level branch on the right side of Figure~\ref{fig:deformcorr}, a multi-head attention layer is employed to aggregate the sampled point features. Unlike the original deformable attention layer~\cite{deformabledetr} that predicts the attention weights by linear projection of the object embedding, we adopt the full Q-K dot-product attention with positional encoding. After being processed by the subsequent layers, the object-level features are finally transformed into to the object-level predictions, consisting of the 3D localization score, weight scale, 3D bounding box size, and other optional properties (velocity and attribute). Note that the attention layer is actually not a necessary component for object-level predictions, but rather a byproduct of the deformable point samples whose features can be leveraged with little computation overhead.

\subsection{Loss Functions for Object-Level Predictions}
As in FCOS3D~\cite{fcos3d}, we adopt smooth L1 regression loss for 3D box size and velocity, and cross-entropy classification loss for attribute. Additionally, a binary cross-entropy loss is imposed upon the 3D localization score, with the target $c_\text{tgt}$ defined as a function of the position error:
\begin{align}
    c_\text{tgt} &= \mathit{Score}(\|t^\ast_\text{XZ} - {t_\text{XZ}}_\text{gt}\|) \notag\\
    &= \max(0, \min(1, -a \log{\|t^\ast_\text{XZ} - {t_\text{XZ}}_\text{gt}\|} + b)),
    \label{eqn:score}
\end{align}
where $t^\ast_\text{XZ}$ is the XZ components of the PnP solution, ${t_\text{XZ}}_\text{gt}$ is the XZ components of the true pose, and $a, b$ are the linear coefficients. The predicted 3D localization score $c_\text{pred}$ shall reflect the positional uncertainty of an object, as a faster alternative to evaluating the uncertainty via the Monte Carlo method during inference (Section~\ref{uncertainpose}). The final detection score is defined as the product of the predicted 3D score and the classification score from the base detector. 

\subsection{Auxiliary Loss Functions}

\begin{figure}[t]
% \vspace{-2ex}
\begin{center}
    \includegraphics[width=1.0\linewidth]{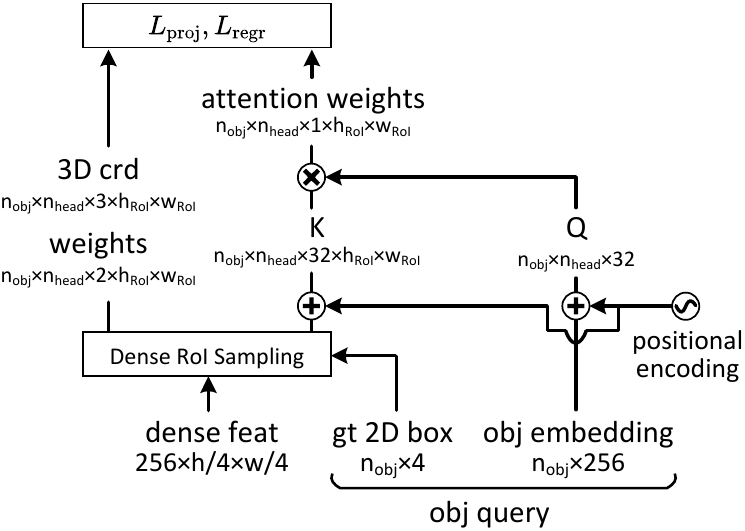}
\end{center}
\vspace{-1ex}
\caption{Architecture of the auxiliary branch. This branch shares the same weights of Q, K projection with the deformable attention layer in the lower right of Figure~\ref{fig:deformcorr}.}
\label{fig:auxiliary}
\end{figure}

To regularize the dense features, we append an auxiliary branch that predicts the multi-head dense 3D coordinates and corresponding weights, as shown in Figure~\ref{fig:auxiliary}. Leveraging the ground truth of object 2D boxes, the features within the box regions are densely sampled via RoI Align~\cite{maskrcnn}, and transformed into the 3D coordinates $x^\text{3D}$ and weights $w^\text{2D}$ via an independent linear layer. Besides, the attention weights $\phi$ are obtained via Q-K dot-product and normalized along the $n_\text{head}$ dimension and across the overlapping region of multiple RoIs via Softmax. 

During training, we impose the reprojection-based auxiliary loss for the multi-head dense predictions, formulated as the negative log-likelihood (NLL) of the Gaussian mixture model~\cite{Bishop94mixturedensity}. The loss function for each sampled point is defined as:
\vspace{-1ex}
\begin{equation}
    L_\text{proj} = \smash[b]{-\log \sum_{\text{RoI}}\sum_{k=1}^{n_\text{head}} \phi_k |\diag{w_k^\text{2D}}| \exp{-\frac{1}{2}\|f_k(y_\text{gt})\|^2}} ,
    \vspace{1mm}
    \label{auxreproj}
\end{equation}
where $k$ is the head index, $f_k(y_\text{gt})$ is the weighted reprojection error of the $k$-th head at the truth pose $y_\text{gt}$. In the above equation, the diagonal matrix $\diag{w_k^\text{2D}}$ is interpreted as the inverse square root of the covariance matrix of the normal distribution, \ie, $\diag{w_k^\text{2D}} = \Sigma^{-\frac{1}{2}}$, and the head attention weight $\phi_k$ is interpreted as the mixture component weight. $\sum_{\text{RoI}}$ is a special operation that takes the overlapping region of multiple RoIs into account, formulating a mixture of multiple heads and multiple RoIs (see code for details).

Another auxiliary loss is the coordinate regression loss that introduces the geometric knowledge. Following MonoRUn~\cite{monorun}, we extract the sparse ground truth of 3D coordinates $x^\text{3D}_\text{gt}$ from the 3D LiDAR point cloud. The multi-head coordinate regression loss for each sampled point with available ground truth is defined as:
\begin{equation}
    L_\text{regr} = \sum_{k=1}^{n_\text{head}} \phi_k \rho\left(\left\| x^\text{3D}_k - x^\text{3D}_\text{gt} \right\|^2\right),
\end{equation}
where $\rho(\cdot)$ is the Huber kernel. $L_\text{regr}$ is essentially a weighted smooth L1 loss (although we write the Huber kernel for convenience in notation).

\begin{table*}[t]
    \RawFloats
    % \vspace*{-2mm}
    \begin{center}
    \scalebox{0.8}{%
    \setlength{\tabcolsep}{0.5em}
    \begin{tabular}{clclcccccc}
        \toprule
        \multirow{2}[2]{*}{ID} &
        \multirow{2}[2]{*}{Method} & \multirow{2}[2]{*}{Data} & \multicolumn{1}{c}{\multirow{2}[2]{*}{NDS}} & \multirow{2}[2]{*}{mAP} & \multicolumn{5}{c}{True positive metrics (lower is better)} \\
        \cmidrule(lr){6-10}
        {} & {} & {} & {} & {} & mATE & mASE & mAOE & mAVE & mAAE \\
        \midrule
        A0 & Basic EPro-PnP & Val & 0.425 & 0.349 & 0.676 & 0.263 & 0.363 & 1.035 & 0.196 \\
        A1 & A0 + coord. regr. & Val & 0.430 & 0.352 & 0.667 & 0.258 & 0.337 & 1.031 & 0.193 \\
        \midrule
        B0 & A0 \textrightarrow\,No reprojection $L_\text{proj}$ & Val & 0.408 & 0.337 & 0.721 & 0.267 & 0.452 & 1.113 & 0.166 \\
        \midrule
        C0 & A0 \textrightarrow\,50\% Monte Carlo score & Val & 0.424 & 0.350 & 0.673 & 0.264 & 0.373 & 1.042 & 0.198 \\
        C1 & A0 \textrightarrow\,100\% Monte Carlo score & Val & 0.424 & 0.350 & 0.675 & 0.264 & 0.367 & 1.048 & 0.199 \\
        \midrule
        D0 & A1 \textrightarrow\,Compact network & Val & 0.434 & 0.358 & 0.672 & 0.264 & 0.351 & 0.983 & 0.181 \\
        D1 & D0 + TTA & Val & 0.446 & 0.367 & 0.664 & 0.260 & 0.320 & 0.951 & 0.179 \\
        \bottomrule
    \end{tabular}}
    \end{center}
    \vspace{-1ex}
    \caption{Additional results of the deformable correspondence network tested on the nuScenes~\cite{nuscenes} benchmark.} 
    \label{tab:addexp}
\end{table*}

\begin{figure}[t]
\vspace{-0.5ex}
\begin{center}
    \includegraphics[width=0.83\linewidth]{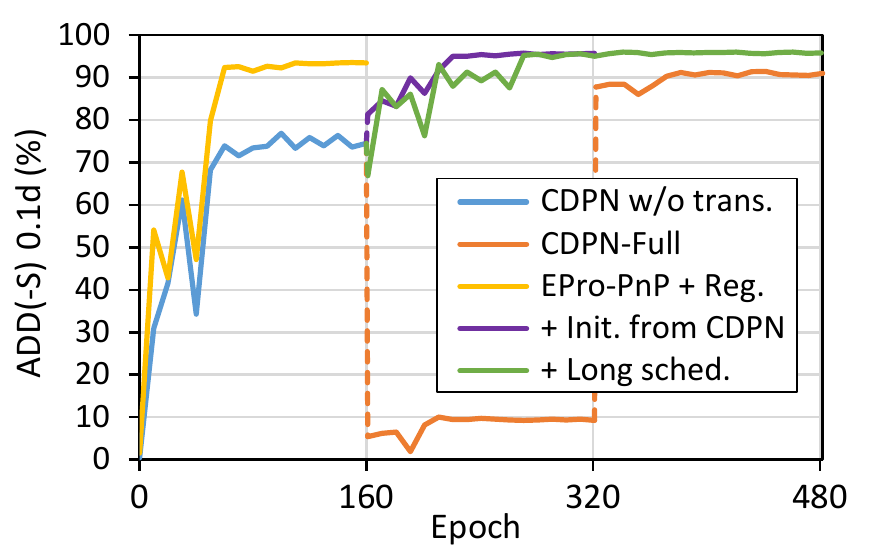}
\end{center}
\vspace{-1.2ex}
\caption{Testing accuracy \vs training progress on LineMOD.}
\label{fig:convergence}
\end{figure}

\subsection{Training Strategy}

During training, we randomly sample 48 positive object queries from the FCOS3D~\cite{fcos3d} detector for each image, which limits the batch size of the deformable correspondence network to control the computation overhead of the Monte Carlo pose loss.

\section{Additional Results of the Dense Correspondence Network}

\subsection{Convergence Behavior}

The convergence behaviors of EPro-PnP and CDPN~\cite{CDPN} are compared in Figure~\ref{fig:convergence}. The original CDPN-Full is trained in 3 stages (rotation head -- translation head -- both together) with a total of 480 epochs. In contrast, EPro-PnP with derivative regularization clearly outperforms CDPN-Full within one stage, and goes further when initialized from the pretrained first-stage CDPN.

\subsection{Inference Time}

Compared to the inference pipeline of CDPN-Full~\cite{CDPN}, EPro-PnP does not use the RANSAC algorithm or extra translation head, so the overall inference speed is more than twice as fast as CDPN-Full (at a batch size of 32), even though we introduces the iterative LM solver. 

Regarding the LM solver itself, inference takes 7.3 ms for a batch of one object, measured on RTX 2080 Ti GPU, excluding EPnP~\cite{EPnP} initialization.
% and Xeon Platinum 8163 2.5 GHz CPU.
As a reference, the state-of-the-art pose refiner RePOSE~\cite{repose} (also based on the LM algorithm) adds 10.9 ms overhead to the base pose estimator PVNet~\cite{pvnet} at the same batch size, measured on RTX 2080 Super GPU,
% and Ryzen 7 3700X 3.6 GHz CPU, 
which is slower than ours.
% \footnote{The speed of our Python-based PnP solver is sensitive to the single-thread performance of the CPU, where the Xeon server processor has the disadvantage.}
Nevertheless, faster inference is possible if the number of points $N=64\times64$ is reduced to an optimal level.

\section{Additional Experiments on the Deformable Correspondence Network}

\subsection{On the Auxiliary Reprojection Loss}
As shown in Table~\ref{tab:addexp}, removing the auxiliary reprojection loss in Eq.~\ref{auxreproj} lowers the 3D object detection accuracy (NDS 0.408 \vs 0.425). Among the true positive metrics, the orientation metric mAOE is the most affected. The results indicate that, although the deformable correspondences can be learned solely with the end-to-end loss, it is still beneficial to add auxiliary task for further regularization, even if the task itself does not involve extra annotation.

\subsection{On the Uncertainty of Object Pose}
\label{uncertainpose}
The dispersion of the inferred pose distribution reflects the aleatoric uncertainty of the predicted pose. Previous work~\cite{monorun} reasons the pose uncertainty by propagating the reprojection uncertainty learned from a surrogate loss through the PnP operation, but that uncertainty requires calibration and is not reliable enough. In our work, the pose uncertainty is learned with the KL-divergence-based pose loss in an end-to-end manner, which is much more reliable in theory. 

To quantitatively evaluate the reliability of the pose uncertainty in terms of measuring the localization confidence, a straightforward approach is to compute the 3D localization score $c_\text{MC}$ via Monte Carlo pose sampling, and compare the resulting mAP against the standard implementation with 3D score $c_\text{pred}$ predicted from the object-level branch. With the PnP solution $t^\ast$, the sampled translation vector $t_j$, and its importance weight $v_j$, the Monte Carlo score is computed by:
\begin{equation}
    c_\text{MC} = \frac{1}{\sum_j v_j} \sum_j v_j \mathit{Score}\left(\|t^\ast_\text{XZ} - {t_\text{XZ}}_j\|\right),
\end{equation}
where the subscript $(\cdot)_\text{XZ}$ denotes taking the XZ components, and the function $\mathit{Score}(\cdot)$ is the same as in Eq.~\ref{eqn:score}. Furthermore, the final score can also be a mixture of the two sources, defined as:
\begin{equation}
    c_\text{mix} = c_\text{MC}^\alpha c_\text{pred}^{1-\alpha},
\end{equation}
where $\alpha$ is the mixture weight.

The evaluation results under different mixture weights are presented in Table~\ref{tab:addexp}. Regarding the mAP metric, the Monte Carlo score is on par with the standard implementation (0.350 \vs 0.350 \vs 0.349), indicating that the pose uncertainty is a reliable measure of the detection confidence. Nevertheless, due to the much longer runtime of inferring with Monte Carlo pose sampling, training a standard score branch is still a more practical choice.

\subsection{On the Network Redundancy and Potential for Future Improvement}
Since the main concern of this paper is to propose a novel differentiable PnP layer, we did not have enough time and resources to fine-tune the architecture and parameters of the deformable correspondence network at the time of submitting the manuscript. Therefore, the network described in Sections~\ref{deformnetmain} and \ref{deformnetsup} was crafted with some redundancy in mind, being not very efficient in terms of FLOP count, memory footprint and inference time, leaving large potential for improvement.

% For example, the dense feature map with a stride of 4 is quite redundant for the 1600\texttimes900 high resolution images of the nuScenes dataset. Besides, we have set the number of points per head $n_\text{hpts}$ to 32, totaling 512 points per object, which is an overkill that affects the efficiency of Monte Carlo sampling. 

To demonstrate the potential for improvement, we train a more compact network with lower resolution (\texttt{stride=8}) for the dense feature map, and the number of points per head $n_\text{hpts}$ reduced from 32 to 16, and squeeze the batch of 12 images into 2 RTX 3090 GPUs. As shown in Table~\ref{tab:addexp}, the overall performance is actually slightly better than the original version (NDS 0.434 \vs 0.430). Still, a more efficient architecture is yet to be determined in future work.

\paragraph{Inference Time}
Regarding the compact network, the average inference time per frame (comprising a batch of 6 surrounding 1600\texttimes672\footnote{The original size is 1600\texttimes900. We crop the images for efficiency.} images, without TTA) is shown in Table~\ref{tab:runtime}, measured on RTX 3090 GPU and Core i9-10920X CPU. On average, the batch PnP solver processes 625.97 objects per frame before non-maximum suppression (NMS).

\begin{table}[h]
    \RawFloats
    \begin{center}
    \scalebox{0.8}{%
    \setlength{\tabcolsep}{0.0em}
    \begin{tabular}{@{\hskip 0.28cm}p{2.5cm} >{\centering\arraybackslash}p{1.4cm} >{\centering\arraybackslash}p{1.4cm} >{\centering\arraybackslash}p{1.4cm} >{\centering\arraybackslash}p{1.4cm} >{\centering\arraybackslash}p{1.4cm}}
        \toprule
        \multirow{2}[2]{*}{PyTorch} & \multirow{2}[2]{*}{\makecell{Backbone \\ \& FPN}} & \multicolumn{2}{c}{Heads} & 
        \multirow{2}[2]{*}{PnP} & 
        \multirow{2}[2]{*}{Total} \\
        \cmidrule(lr){3-4}
        {} & {} & FCOS & Deform \\
        \midrule
        v1.8.1+cu111 & 0.195 & 0.074 & 0.028 & \textbf{0.026} & 0.327 \\
        v1.10.1+cu113 & 0.172 & 0.056 & 0.025 & 0.045 &\textbf{ 0.301} \\
        \bottomrule
    \end{tabular}}
    \end{center}
    \vspace{-1ex}
    \caption{Inference time (sec) of the deformable correspondence network on nuScenes object detection dataset~\cite{nuscenes}. The PnP solver (including the random sampling initialization in Section~\ref{rslm}) works faster (26 ms) with PyTorch v1.8.1, for which the code was originally developed, while the full model works faster (301 ms) with PyTorch v1.10.1.}
    \label{tab:runtime}
\end{table}

\section{Limitation}

EPro-PnP is a versatile pose estimator for general problems, yet it has to be acknowledged that training the network with the Monte Carlo pose loss is inevitably \emph{slower} than the baseline. At the batch size of 32, training the CDPN (without translation head) takes 143 seconds per epoch with the original coordinate regression loss, and 241 seconds per epoch with the Monte Carlo pose loss, which is about 70\% longer time, as measured on GTX 1080 Ti GPU. However, the training time can be controlled by adjusting the number of Monte Carlo samples or the number of 2D-3D corresponding points. In this paper, the choice of these hyperparameters generally leans towards redundancy. 

\vfill\null
\newpage

\section{Additional Visualization}

\begin{figure}[h]
%   \vspace{-1mm}
   \begin{center}
   \includegraphics[width=0.95\linewidth]{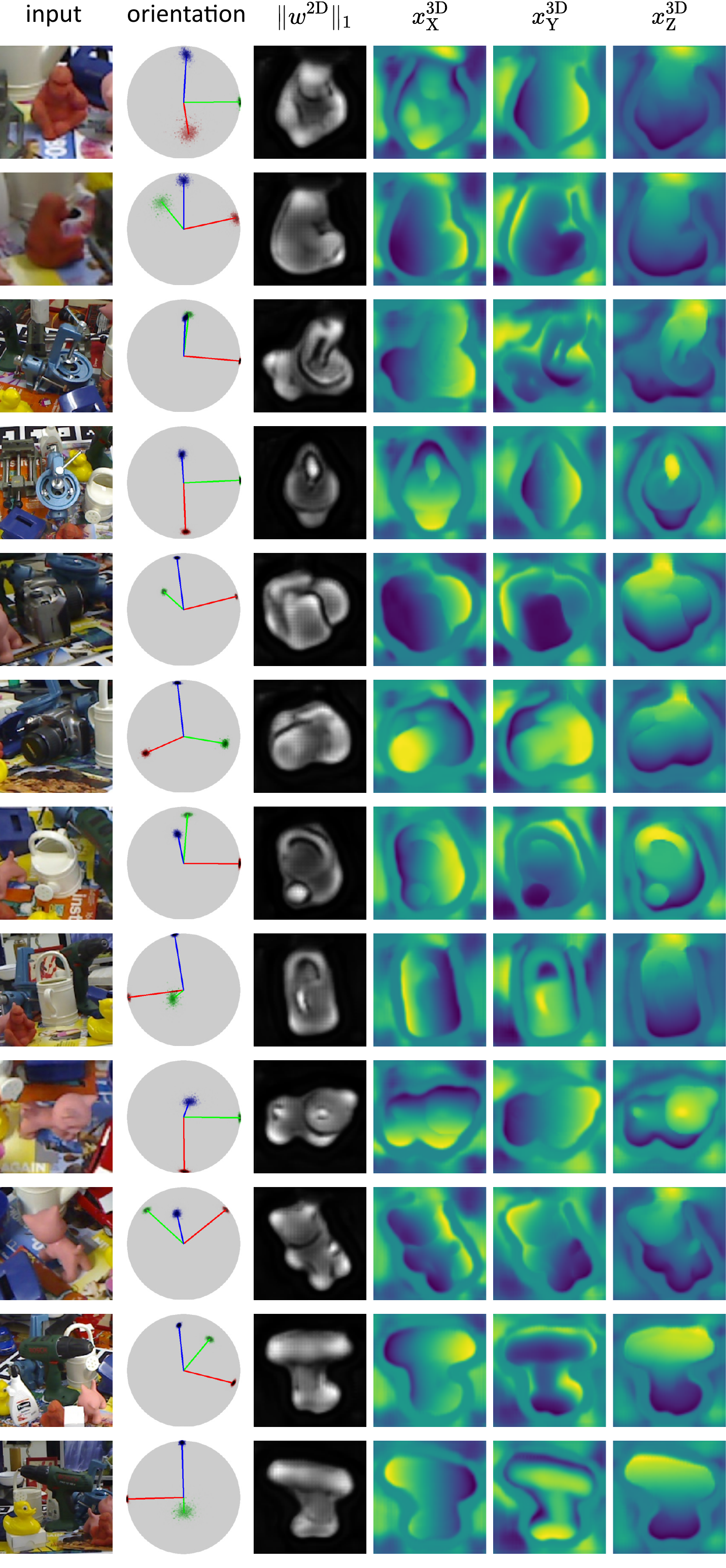}
   \end{center}
   \vspace{-2mm}
   \caption{Inferred results on LineMOD test set by EPro-PnP with derivative regularization and pretrained CDPN weights, Part I.} 
\end{figure}
\vfill\null

\begin{figure}[t]
   \vspace{-2mm}
   \begin{center}
   \includegraphics[width=0.95\linewidth]{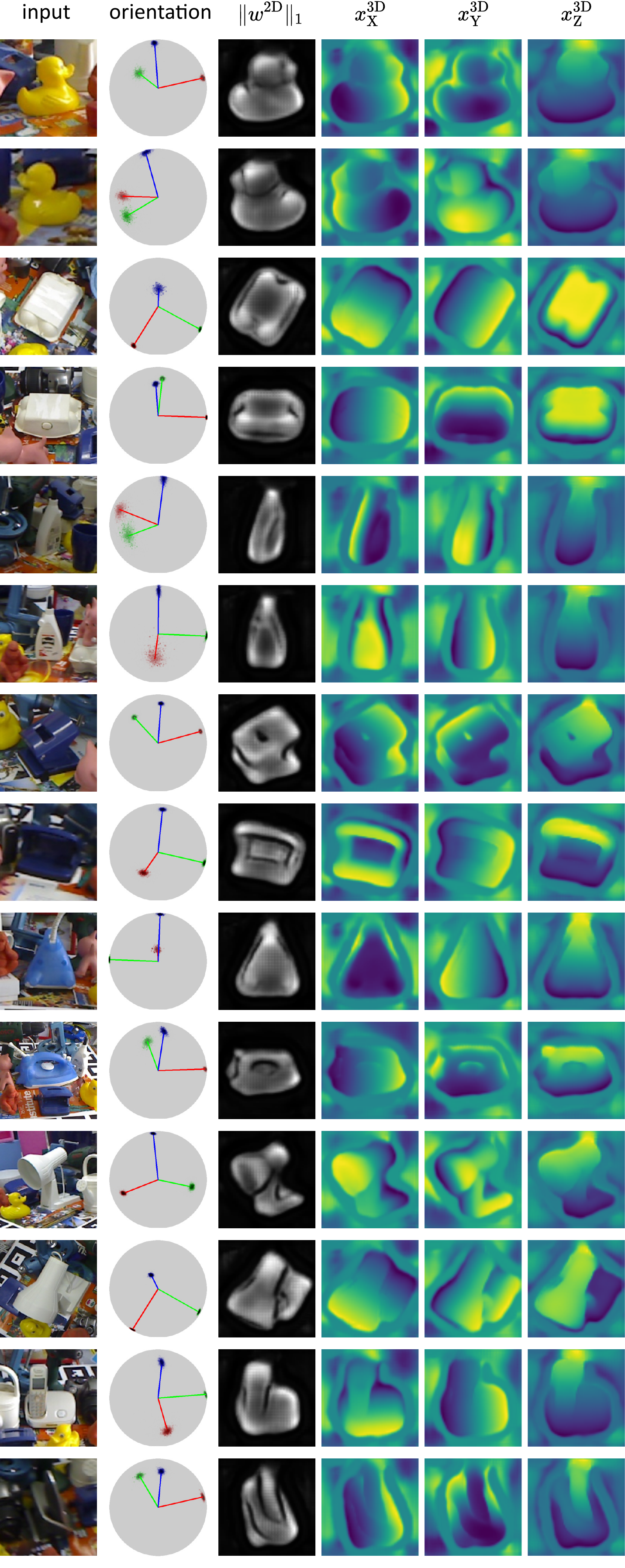}
   \end{center}
   \vspace{-2mm}
   \caption{Inferred results on LineMOD test set by EPro-PnP with derivative regularization and pretrained CDPN weights, Part II.} 
\end{figure}

\begin{figure}[t]
   \begin{center}
   \includegraphics[width=0.8\linewidth]{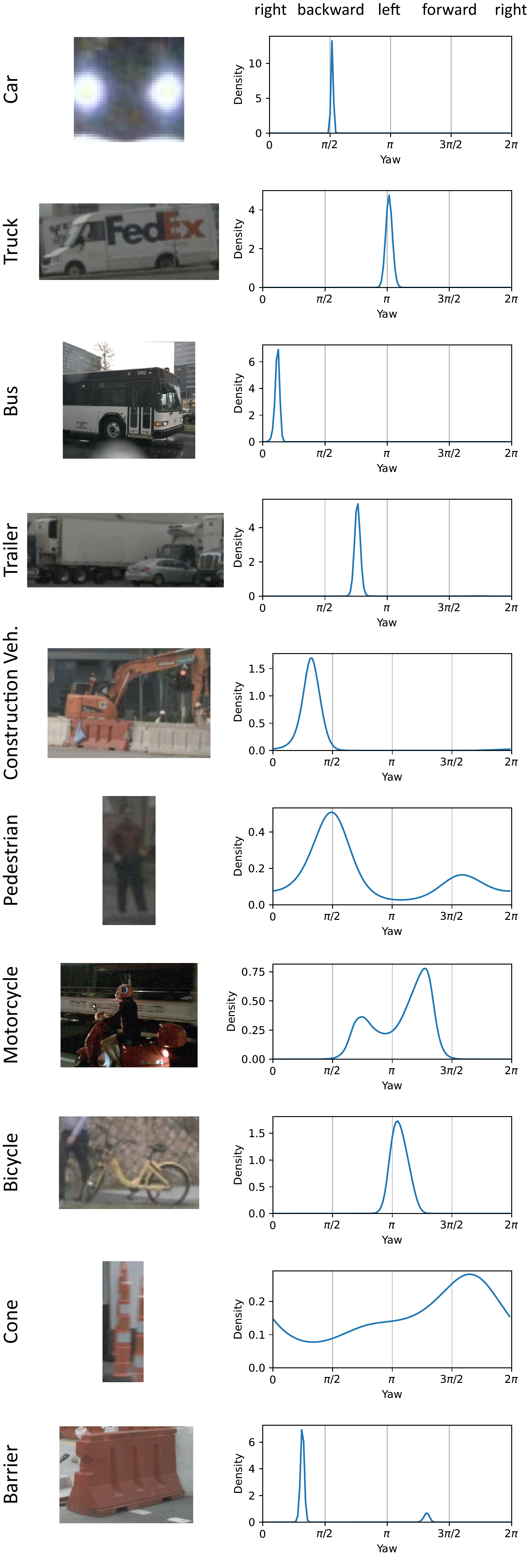}
   \end{center}
   \vspace{-3mm}
   \caption{Inferred orientation on nuScenes validation set by the Basic EPro-PnP.} 
\end{figure}

\begin{figure*}[t]
   \captionsetup{font=normalsize}
   \begin{center}
   \includegraphics[width=0.88\textwidth]{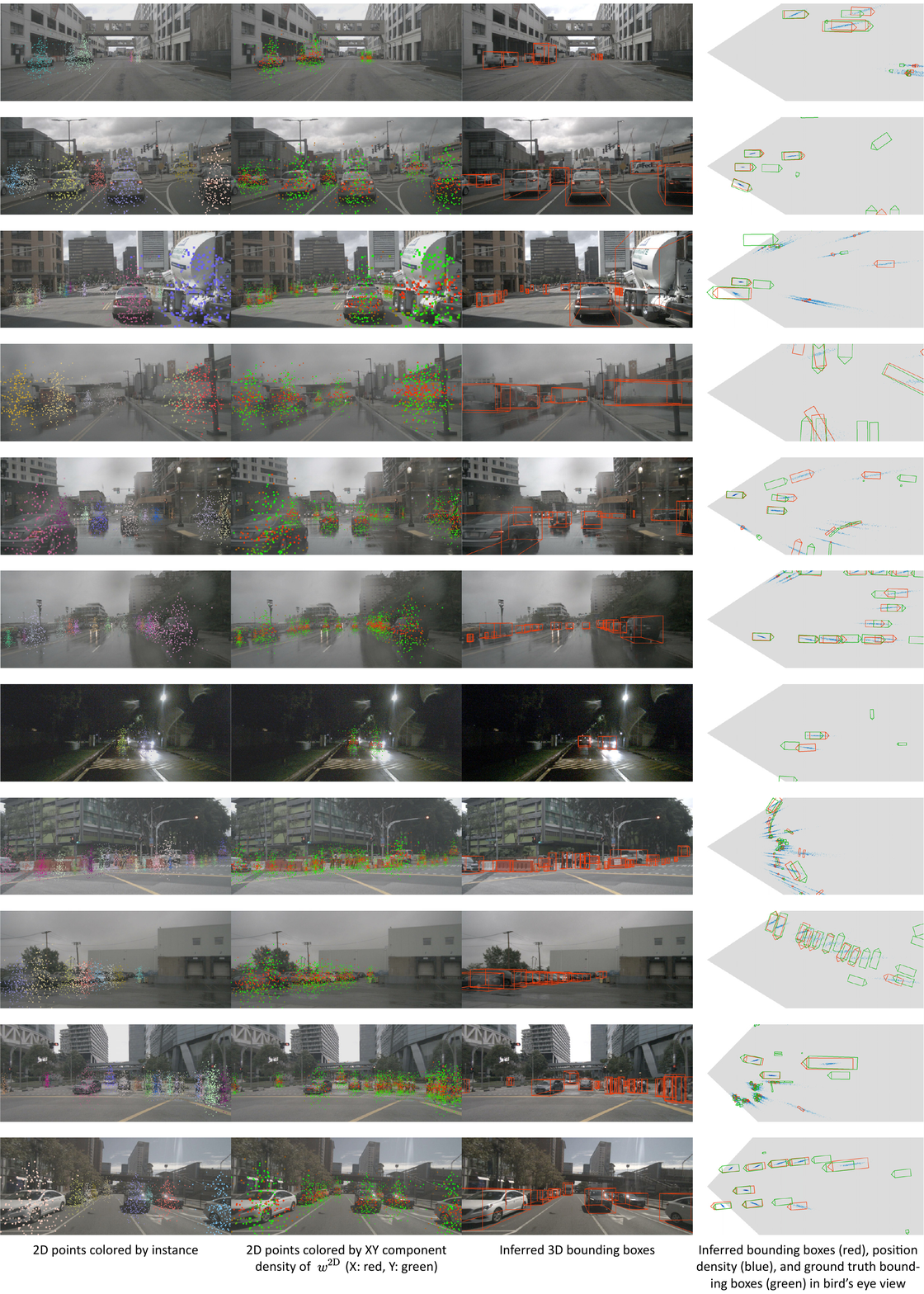}
   \end{center}
   \vspace{-3mm}
   \caption{Inferred results on nuScenes validation set by the Basic EPro-PnP.} 
\end{figure*}

\clearpage
\section{Notation}

\begin{table*}[b]
    \captionsetup{font=normalsize}
    \RawFloats
    \begin{center}
    \scalebox{1.0}{%
    \setlength{\tabcolsep}{0.5em}
    \begin{tabular}{cll}
        \toprule
        \multicolumn{2}{l}{Notation} & Description \\
        \midrule 
        $x^\text{3D}_i$ & $\in \mathbb{R}^3$ & Coordinate vector of the $i$-th 3D object point\\
        $x^\text{2D}_i$ & $\in \mathbb{R}^2$ & Coordinate vector of the $i$-th 2D image point\\
        $w^\text{2D}_i$ & $\in \mathbb{R}^2_+$ & Weight vector of the $i$-th 2D-3D point pair\\
        $X$ & & The set of weighted 2D-3D correspondences\\
        
        $y$ & & Object pose\\
        $y_\text{gt}$ & & Ground truth of object pose\\
        $y^\ast$ & & Object pose estimated by the PnP solver\\
        
        $R$ & & 3\texttimes3 rotation matrix representation of object orientation\\
        $\theta$ & & 1D yaw angle representation of object orientation\\
        % $\theta^\ast$ & & 1D yaw angle estimated by the PnP solver \\
        $l$ & & Unit quaternion representation of object orientation\\
        % $l^\ast$ & & Unit quaternion estimated by the PnP solver\\
        $t$ & $\in \mathbb{R}^3$ & Translation vector representation of object position\\
        $\Sigma_{y^\ast}$ & & Pose covariance estimated by the PnP solver \\
        
        $J$ & & Jacobian matrix\\
        $\tilde{J}$ & & Rescaled Jacobian matrix\\
        $F$ & & Concatenated vector of weighted reprojection errors of all points\\
        $\tilde{F}$ & & Concatenated vector of rescaled weighted reprojection errors of all points\\
        
        $\pi(\cdot)$ & $: \mathbb{R}^3 \rightarrow \mathbb{R}^2 $ & Camera  projection function\\
        $f_i(y)$ & $\in \mathbb{R}^2$ & Weighted reprojection error of the $i$-th correspondence at pose $y$\\
        $r_i(y)$ & $\in \mathbb{R}^2$ & Unweighted reprojection error of the $i$-th correspondence at pose $y$\\
        $\rho(\cdot)$ & & Huber kernel function\\
        $\rho^\prime_i$ & & The derivative of the Huber kernel function of the $i$-th correspondence\\
        $\delta$ & & The Huber threshold\\
        $p(X|y)$ & & Likelihood function of object pose\\
        $p(y)$ & & PDF of the prior pose distribution\\
        $p(y|X)$ & & PDF of the posterior pose distribution\\
        $t(y)$ & & PDF of the target pose distribution\\

        $q(y), q_t(y)$ & & PDF of the proposal pose distribution (of the $t$-th AMIS iteration)\\
        $y_j, y_j^t$ & & The $j$-th random pose sample (of the $t$-th AMIS iteration)\\
        $v_j, v_j^t$ & & Importance weight of the $j$-th pose sample (of the $t$-th AMIS iteration)\\
        
        $i$ & & Index of 2D-3D point pair\\
        $j$ & & Index of random pose sample\\
        $t$ & & Index of AMIS iteration\\
        $N$ & & Number of 2D-3D point pairs in total\\
        $K$ & & Number of pose samples in total\\
        $T$ & & Number of AMIS iterations\\
        $K^\prime$ & & Number of pose samples per AMIS iteration\\
        $n_\text{head}$ & & Number of heads in the deformable correspondence network\\
        $n_\text{hpts}$ & & Number of points per head in the deformable correspondence network\\
        $L_\text{KL}$ & & KL divergence loss for object pose\\
        $L_\text{tgt}$ & & The component of $L_\text{KL}$ concerning the reprojection errors at target pose\\
        $L_\text{pred}$ & & The component of $L_\text{KL}$ concerning the reprojection errors over predicted pose\\
        $L_\text{reg}$ & & Derivative regularization loss\\
        % $L_\text{pos}$ & & The position component of $L_\text{reg}$\\
        % $L_\text{orient}$ & & The orientation component of $L_\text{reg}$\\
        \bottomrule
    \end{tabular}}
    \end{center}
    \vspace{-1ex}
    \caption{A summary of frequently used notations.} 
    \vspace*{2.2cm}
    \label{tab:notation}
\end{table*}

\clearpage

\end{document}